\documentclass[runningheads]{llncs}

\usepackage{eccv}

\usepackage{eccvabbrv}
\usepackage{wrapfig}
\usepackage{graphicx}
\usepackage{booktabs}
\usepackage{caption}
\usepackage{xcolor}
\usepackage{ifthen}
\usepackage{textcomp}
\usepackage{color}
\usepackage{tikz}
\usepackage{comment}
\usepackage{amsmath,amssymb} % define this before the line numbering.
\usepackage{microtype}
\usepackage{enumitem}
\usepackage{makecell}
\usepackage{multirow}
\usepackage{caption}
 \usepackage{lipsum}

\usepackage[accsupp]{axessibility}  % Improves PDF readability for those with disabilities.

\usepackage{adjustbox}
\newcommand{\Gen}{\mathcal{G}}
\newcommand{\MLP}{\mathcal{M}}
\newcommand{\Interp}{\psi}

\newcommand{\Detach}{\mathcal{D}}

\newcommand{\token}{t_{\text{init}}}
\newcommand{\tokenXY}{t_{XY}}

\usepackage[accsupp]{axessibility}  % Improves PDF readability for those with disabilities.

\usepackage{hyperref}

\usepackage{orcidlink}
\usepackage{marvosym}
 \usepackage{lipsum}

\newcommand\blfootnote[1]{%
  \begingroup
  \renewcommand\thefootnote{}\footnote{#1}%
  \addtocounter{footnote}{-1}%
  \endgroup
}
 
\begin{document}

% \title{Robust View Synthesis with Triplane Generator and Pose oriented feature Aggregation} 

% \title{xxxxx : Hybrid Triplane Represents Radiance Field with Noisy Posed Images} 
% \title{Robust-Triplane: Disentangled Generation and Aggregation for Robust Radiance Fields} 
\title{Disentangled Generation and Aggregation \\ for Robust Radiance Fields}

%  : Hybrid Triplane Represents Radiance Field for 

% Robust View Synthesis with Triplane Generator and Pose oriented feature Aggregation
% \title{Triplane Generator Meets Noisy Posed Images} 


\author{Shihe Shen \inst{1}\textsuperscript{$\star$} \and
Huachen Gao\inst{1}\textsuperscript{$\star$} \and Wangze Xu \inst{1} \and Rui Peng \inst{1,2} \and 
\\ Luyang Tang \inst{1,2} \and Kaiqiang Xiong \inst{1,2} \and Jianbo Jiao \inst{3} \and Ronggang Wang\inst{1,2,}\textsuperscript{\Letter} }

\authorrunning{S. Shen, H. Gao et al.}

\institute{School of Electronic and Computer Engineering, Peking University \and Peng Cheng Laboratory \and School of Computer Science, University of Birmingham\\
\email{\{shshen0308, gaohuachen712\}@gmail.com rgwang@pkusz.edu.cn}
\blfootnote{\textsuperscript{$\star$} Equal contribution.}
}

\maketitle

\begin{abstract}
% The abstract should summarize the contents of the paper. 
% LNCS guidelines indicate it should be at least 70 and at most 150 words.
% Please include keywords as in the example below. 
% This is required for papers in LNCS proceedings.
% \keywords{First keyword \and Second keyword \and Third keyword}

The utilization of the triplane-based radiance fields has gained attention in recent years due to its ability to effectively disentangle 3D scenes with a high-quality representation and low computation cost. A key requirement of this method is the precise input of camera poses. However, due to the local update property of the triplane, a similar joint estimation as previous joint pose-NeRF optimization works easily results in local minima. To this end, we propose the Disentangled Triplane Generation module to introduce global feature context and smoothness into triplane learning, which mitigates errors caused by local updating. Then, we propose the Disentangled Plane Aggregation to mitigate the entanglement caused by the common triplane feature aggregation during camera pose updating. In addition, we introduce a two-stage warm-start training strategy to reduce the implicit constraints caused by the triplane generator. Quantitative and qualitative results demonstrate that our proposed method achieves state-of-the-art performance in novel view synthesis with noisy or unknown camera poses, as well as efficient convergence of optimization. Project page: \href{https://gaohchen.github.io/DiGARR/}{https://gaohchen.github.io/DiGARR/}.
\end{abstract}

% \vspace{-0.5cm}
\newcommand{\Tensor}{\mathcal{T}}
\newcommand{\Vector}{\mathbf{v}}
\newcommand{\Matrix}{\mathbf{M}}
\newcommand{\AppVec}{\mathbf{b}}
\newcommand{\AppMat}{\mathbf{B}}
\newcommand{\VectorPack}{\mathbf{V}}
\newcommand{\RRR}{\mathbb{R}}
\newcommand{\DimIJK}{I\times J\times K}
\newcommand{\DimCh}{P}

\newcommand{\OuterP}{\circ}
\newcommand{\ScalarP}{\ast}
\newcommand{\Pos}{\mathbf{x}}
\newcommand{\Dens}{\sigma}
\newcommand{\Rad}{c}
\newcommand{\Color}{C}
\newcommand{\Trans}{\tau}
\newcommand{\Step}{\Delta}
\newcommand{\Comp}{\mathcal{A}}
\newcommand{\Dir}{d}
\newcommand{\Grid}{\mathcal{G}}
\newcommand{\ShadFunc}{S}
\newcommand{\loss}{\mathcal{L}}

\begin{figure}[t]
    \includegraphics[width=\linewidth]{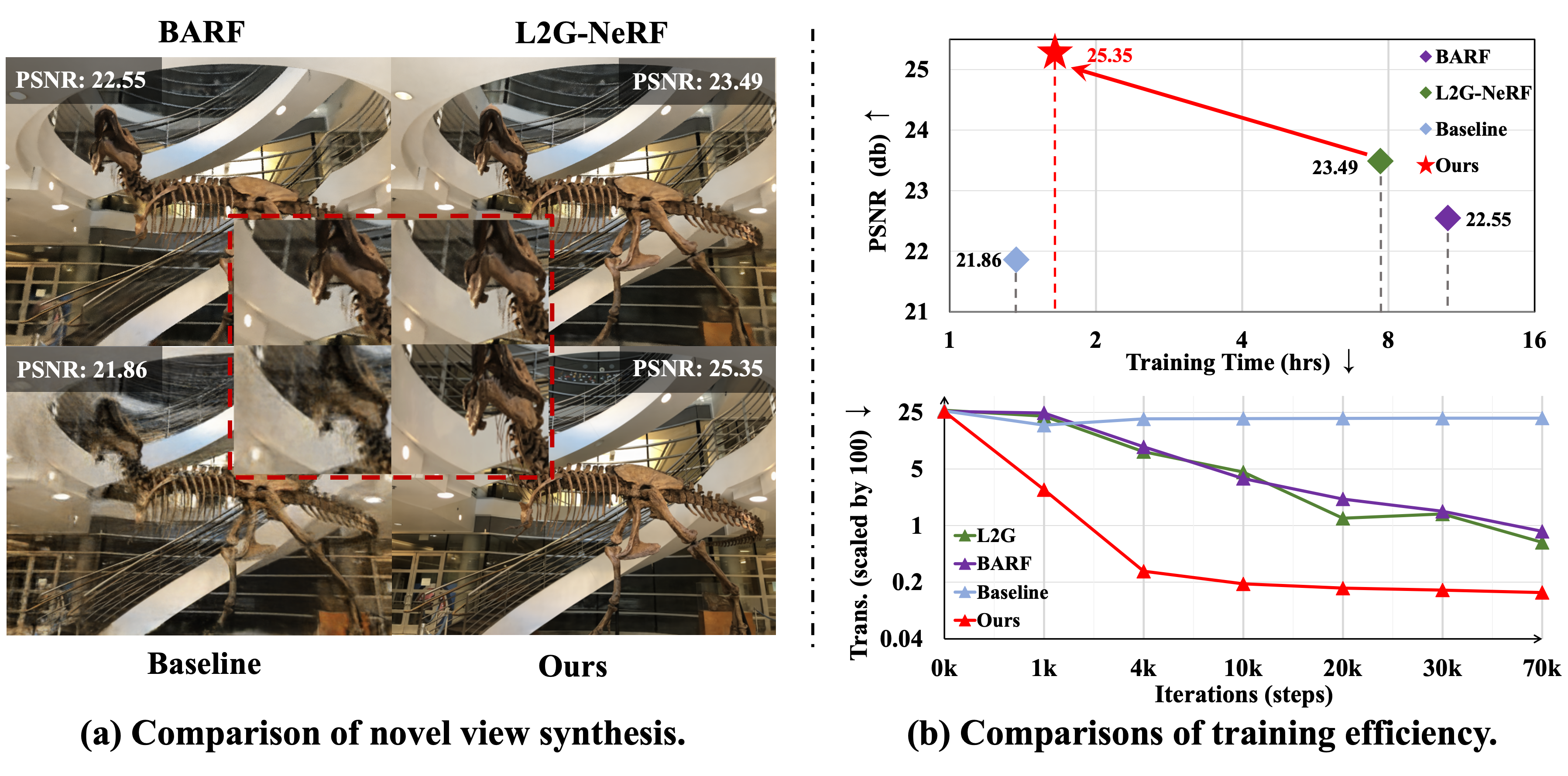}
    \centering
    % \vspace{-6mm}
    \caption{
    \textbf{Comparisons of novel view synthesis and training efficiency on \textit{Trex}.} (a) shows novel view synthesis results of different works, where the baseline is the direct joint pose-triplane optimization; (b) shows efficiency comparisons where the upper shows training time and PSNR, and the lower shows iterations and translation errors. 
    } 
    \label{fig:teaser}
    % \vspace{-0.8cm}
\end{figure}

\section{Introduction}
\label{sec:intro}

Recently, adopting neural networks has become increasingly popular for Novel View Synthesis (NVS), where the Neural Radiance Fields (NeRF) \cite{mildenhall2021nerf} has brought a great surge in high-fidelity synthesis quality. NeRF represents a 3D radiance field by multi-layer perceptrons (MLPs), and renders novel views by differentiable volume rendering \cite{rendering1}. A crucial prerequisite for achieving promising rendering results with NeRF is the precise annotated camera parameters. However, accurate camera poses are not easily attainable, which heavily relies on external Structure-from-Motion (SfM) algorithms like COLMAP \cite{colmap}.

% 加几个引用
To this end, there has been a growing focus on methods for reducing the reliance on camera parameters by optimizing NeRF and camera parameters simultaneously \cite{wang2021nerfmm,lin2021barf,chng2022garf,nopenerf,l2gnerf}. 
However, based on the vanilla NeRF-MLP representation, most existing approaches demand hours even days for training with modern powerful GPUs, severely limiting their use in practical applications.
Therefore, it is significant to improve training efficiency.
An intuitive idea for acceleration is to explore alternative representations for MLP, such as voxels\cite{dvgo,plenoxels}, hash-grids\cite{instantngp}, triplanes\cite{kplanes,hexplane} and Gaussian splats\cite{3dgs}.
In this paper, we propose a novel triplane-based 3D representation to estimate camera poses and the 3D scene, with both high efficiency and quality.
% Widely applied in many recent works \cite{kplanes,hexplane,tensorf,LRM}, triplane is a highly disentangled explicit representation of 3D scenes, which accelerates pose-NeRF joint optimization.
Triplane is a highly disentangled explicit representation of 3D scenes, which has been widely applied in many recent works \cite{kplanes,hexplane,tensorf,LRM}.
With a high data compression ratio, low computational cost and comparable performance, triplane is more concise and scalable than volumes, thus appropriate to accelerating pose-NeRF optimization.
% As a concise and scalable representation, triplane achieves a high data compression ratio, low computational cost, and comparable performance with other representations as volumes.
% compared to other representations such as voxel volumes. 
% Triplane is a direct decoupled representation, which accelerates pose-NeRF joint optimization.
% Compared to explicit representations like volumes, triplane is more compact, efficient, and widely used in concurrent works \cite{kplanes,hexplane,tensorf,LRM}.
Meanwhile, as a structured 3D representation with fixed-size feature maps, triplane is more controllable and suitable for bundle adjusting tasks, while concurrent pose-3DGS joint optimizations \cite{li2024ggrt,cf3dgs} are primarily restricted to video streams or ordered image collection.
% Compared with 3DGS, triplane is a structured 3D representation with fixed-size feature maps and is more controllable and suitable for bundle adjusting tasks.

\begin{figure}[t]
    \includegraphics[width=0.88\linewidth]{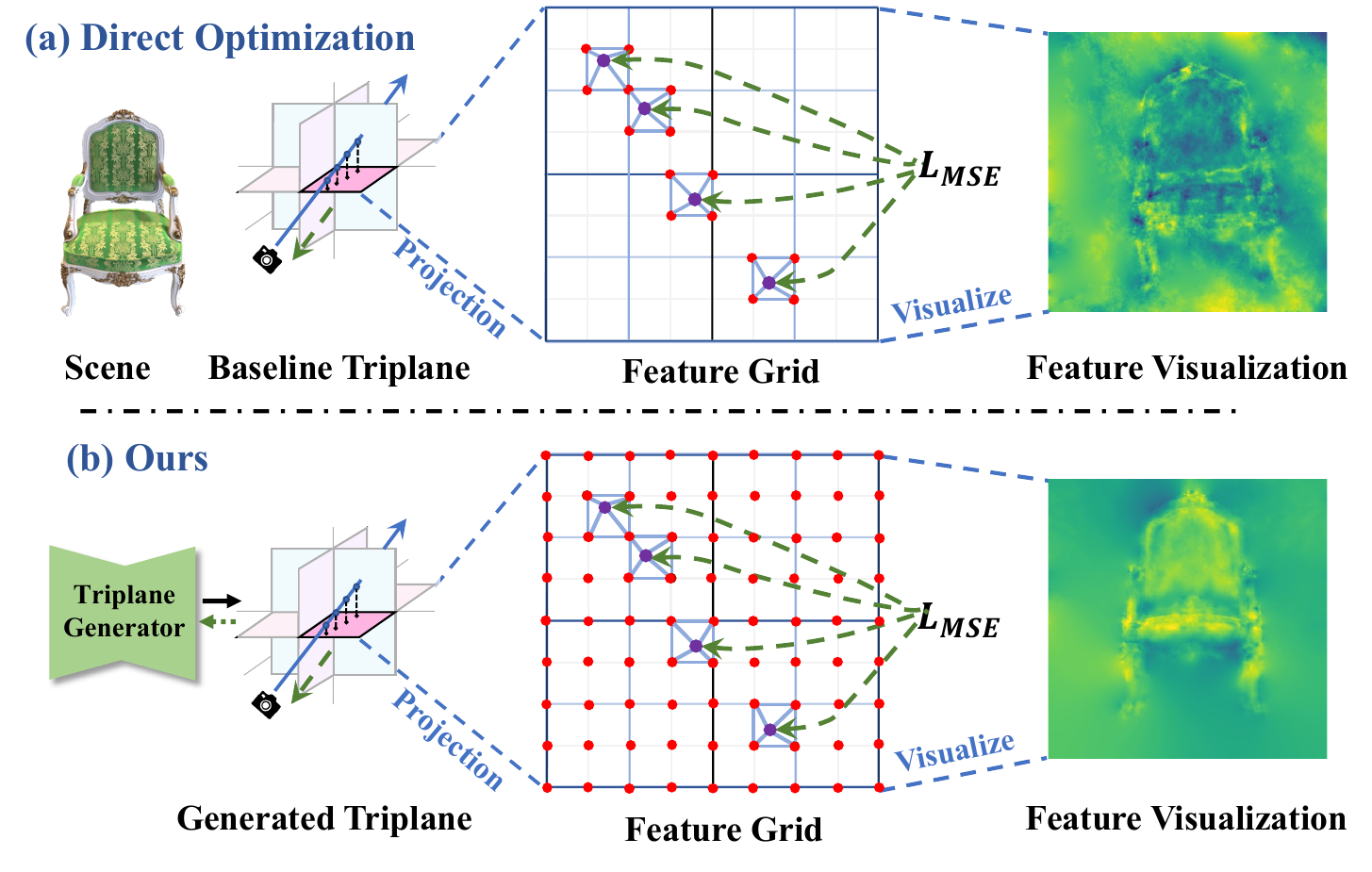}
    \centering
    % \vspace{-5mm}
    % \captionsetup{font=scriptsize}
    \caption{
    \textbf{Illustration of the local updating on feature grids.} Feature points in \textcolor[RGB]{255,0,0}{\textbf{red}} on the planes will be updated during one forward pass, lines in \textcolor[RGB]{56,87,35}{\textbf{green}} represent gradient flow. (a) shows the updated feature points by direct optimization, and (b) shows the updated feature points after adopting the triplane generator.
    }
    \label{fig:local}
    % \vspace{-0.8cm}
\end{figure}

However, introducing a triplane to the joint optimization is nontrivial. We observe that the naive direct combination of pose estimation and triplane radiance field is prone to get trapped into local-minima, especially in the early training stage, and could hardly get rectified, resulting in unsatisfactory view synthesis results (\cref{fig:teaser} (a)). 
The main reason lies in the following two aspects. 
(1) The triplane radiance field follows a local updating policy on each feature grid as shown in \cref{fig:local} (a), since features are derived from interpolation and planes are directly updated only in a few grids used for interpolation. When the camera poses and the triplane are ambiguous, the inaccurate poses lead to ray transmission bias. The local updating property of the triplane incorrectly updates the projected feature of sampled 3D points on biased rays. As the training proceeds, the error from local updating can not be rectified, where the feature parameters on the triplane fail to perceive the global contextual information. This causes additional ambiguity between camera poses and scene reconstructions, finally resulting in local minima of the joint optimization.
(2) As triplane decoupling 3D scene into three orthogonal planes, anisotropic features need to be aggregated among independent planes. With different learning complexities of planes and the introduced entanglement with pose and planes, commonly used aggregations as sum \cite{eg3d} and production \cite{kplanes} impose conflicting signals on pose optimization, thus failing to achieve accurate pose estimation and expressive scene representation simultaneously.

To address the above issues of pose and triplane joint optimization, we propose a triplane-based representation with disentangled generation and aggregation.
Firstly, we obtain triplane features from a disentangled generator, with frozen noise tokens of each plane as input, possessing both global smoothness in implicit representation and training efficiency in explicit representation. As shown in \cref{fig:local} (b), by parameterizing the triplane with triplane generator, network parameters are shared by all grids on the plane, and the information for feature updating will radiate around the plane, introducing global context to eliminate local error accumulating without additional constraints or post-processing steps on triplane.
Secondly, digging into how triplane feature aggregation impacts camera poses during joint optimization, we design a novel Disentangled Plane Aggregation (DPA) to relieve optimizing collision among three feature planes for a single 3D sampled point. The DPA disentangles pose with triplane feature by distributing updating signals to pose and triplane respectively, realizing robust and unambiguous joint optimization of pose and scene representation.
Additionally, the smoothness introduced by the triplane generator has a side effect on high-frequency details of scene representation. To preserve our high-frequency feature and promote scene expressiveness, we present a two-stage warm-start training strategy. Inheriting the relatively accurate poses and a coarse triplane along with the MLP decoder from the first-stage training, we seamlessly transform to a direct feature optimization joint with pose refinement in the second stage.

In summary, the primary contributions are as follows:
\begin{itemize}
% \setlength{\itemsep}{2pt}
% \vspace{-0.1cm}
\setlength{\parsep}{0pt}
\setlength{\parskip}{0pt}
    \item We present a hybrid disentangled scene representation based on triplane for joint estimation of camera poses and novel view synthesis. 
    
    \item We leverage the triplane generator for joint estimation to address local errors in feature grids caused by the local updating in baseline pose-triplane optimization. To the best of our knowledge, our method is the first attempt to incorporate the deep network prior as a hybrid representation in the field of joint pose-NeRF optimization.

    \item We analyze the effect of different aggregation approaches on the pose estimation, and design a new disentangled plane aggregation method. We also introduce a two-stage warm-start training strategy to mitigate the smoothing on triplane grids caused by the triplane generator.

    \item Qualitative and quantitative results on real-world LLFF \cite{llff} and NeRF-synthetic \cite{mildenhall2021nerf} dataset show that our method achieves state-of-the-art performance on both novel view synthesis and pose estimation.
\end{itemize}

\section{Related Work}

% \vspace{-0.2cm}
\subsubsection{Novel View Synthesis.}
NeRF \cite{mildenhall2021nerf} has become a popular representation in the field of novel view synthesis for its high-fidelity rendering results. Many follow-up works are proposed to improve NeRF's performance. \cite{mipnerf, trimiprf} replace the ray casting with anti-aliased cone tracing, and \cite{mipnerf360} is further proposed to handle the unbounded scene with non-linear scene parameterization.\cite{zhang2020nerf++} separates foreground and background using proposed sampling algorithms, \cite{infonerf,regnerf,simplenerf} uses additional constraints, while \cite{ddpnerf,dsnerf, wei2021nerfingmvs,darf} adopts depth priors to improve NeRF's geometry learning. NeRF has also gained many vision or graphic applications, such as surface reconstruction \cite{neus,unisurf,gens}, dynamic scene reconstruction \cite{hypernerf, nerfies, dynerf, d-nerf, hexplane, kplanes}, and single view reconstruction \cite{sinnerf, zero123,LRM,SSDNeRF}. Most recently, 3D Gaussian Splatting \cite{3dgs} also demonstrated strong capability in novel view synthesis.

To address the slow rendering speed of the vanilla MLP-based NeRF, many efforts tried to apply explicit grid-based representations, such as multi-resolution hash encoding \cite{instantngp,zipnerf}, voxel grids \cite{dvgo,plenoxels,plenoctree,nsvf} and triplanes \cite{eg3d,tensorf,kplanes,trimiprf}. The triplane representation leads to faster optimization and more compact modeling than other representations, which is widely adopted in recent works \cite{LRM,PF-LRM,triplanegaussian}.

% \vspace{-0.3cm}
\subsubsection{Joint Pose and NeRF Optimization.}
Pose-NeRF joint optimization has been widely studied recently. iNeRF \cite{yen2021inerf} first shows the ability of pose estimation using reconstructed NeRF models. \cite{wang2021nerfmm} first jointly estimates camera intrinsic, extrinsic and NeRF. \cite{lin2021barf} proposed a coarse-to-fine positional encoding method. \cite{scnerf} estimated camera distortion and proposed a geometric regularization. \cite{meng2021gnerf,zhang2022vmrf} adopted GANs, while \cite{xia2022sinerf,chng2022garf} modified different activations for more suitable pose estimation. Recently, \cite{nopenerf, fdcnerf} utilized mono-depth estimation \cite{pdanet,unimvsnet} for geometry guidance, which is designed for long-sequence images. \cite{l2gnerf} proposed a local-to-global registration to estimate poses from learned multiple local poses. \cite{sparf} used pre-computed correspondences as priors for sparse view settings. \cite{melon,LU-NeRF} optimized NeRF without any pose initialization. \cite{localrf,robustdynerf} are proposed for the long sequence static and dynamic videos respectively. \cite{robust_hash} accelerated previous pose-NeRF joint optimization using multi-resolution hash encoding. In this paper, we propose to use a fast and compact triplane for joint estimation.
% \vspace{-0.1cm}
\section{Preliminaries}
\label{sec:preliminary}
In this section, we first introduce the common pipeline of pose and NeRF joint learning, then we present the pipeline of our baseline directly optimization of triplane with noisy or unknown pose prior (denote as baseline).
% \vspace{-0.2cm}
\subsubsection{Formulation of Pose-NeRF Optimization.}
Given a set of images $\{I_i\}_{i=1}^N$ with camera intrinsics and noisy or \textit{unknown} extrinsics $\{\mathbf{T}_i\}_{i=1}^N$, our goal aims to reconstruct triplane-based radiance field and estimate the accurate camera poses $\mathbf{T}$. Denote a sampled 3D point $\mathbf{x} \in \mathbb{R}^3$ along a camera ray emitted from camera $i$, the color $\mathbf{c}$ and density $\mathbf{\sigma}$ at $\mathbf{x}$ can be derived from a MLP-based NeRF $f:\mathbb{R}^3\rightarrow\mathbb{R}^4$ as $\mathbf{c}, \sigma = f(\mathbf{x};\mathbf{\Theta})$.
A synthesized image $\hat I$ can be rendered along camera rays $\mathbf{r}(t)=\mathbf{o}+t\mathbf{d}$ between near and far plane $t_{n}$ and $t_{f}$. The volume rendering function $\hat{\mathcal{I}}$ \cite{rendering1,rendering2} is formulated as:
% \vspace{-0.2cm}
\begin{equation}
\label{eqn:render}
\hat{\mathcal{I}}(\mathbf{r})=\int_{t_n}^{t_f} W(t) \sigma(\mathbf{r}(t)) \mathbf{c}(\mathbf{r}(t), \mathbf{d}) \ \mathrm{d} t, 
% \vspace{-0.2cm}
\end{equation}
where $W(t)=exp(-\int_{t_n}^{t}\sigma(\mathbf{r}(s)) \ \mathrm{d} s)$ is accumulated transmittance along the ray. The photometric loss is $\mathcal{L}_{Render}=\Sigma_{i=1}^N ||I_i - \hat{I}_i ||_2^2.$
Taking the photometric loss as total loss $\mathcal{L}$, triplane NeRF parameters $\mathbf{\Theta}$ and camera poses $\mathbf{T}$ which are used to cast the ray can be optimized by minimizing the total loss:
% \vspace{-0.1cm}
\begin{equation}
    \mathbf{T}^*, \mathbf{\Theta}^* = \mathop{argmin} \limits_{\mathbf{T},\mathbf{\Theta}}\mathcal{L}(
    \hat{\mathbf{T}}, \hat I|I),
% \vspace{-0.4cm}
\end{equation}
where $\hat{\mathbf{T}}$ denotes that learnable camera parameters are updated during optimization, and $\mathbf{T}^*, \mathbf{\Theta}^*$ are the final optimized parameters.
% \vspace{-0.3cm}
\subsubsection{Baseline Joint Optimization of Pose and Triplane.}
In our baseline, the 3D scene can be factorized by the triplane representation which contains three axis-aligned feature planes $\mathcal{P}_{XY}$, $\mathcal{P}_{YZ}$ and  $\mathcal{P}_{XZ}$. The features corresponding to sample $\mathbf{x}$ on plane $k$ can be queried by projection and interpolation:
% \vspace{-0.15cm}
\begin{equation}
    \label{eqn:feat_k}
    F_k = \psi\left(\mathcal{P}_k, \pi_i(\mathcal{N}(\mathbf{x}))\right), k \in \{XY,XZ,YZ\},
    % \vspace{-0.1cm}
\end{equation}
where $\pi_k$ projects $\mathbf{x}$ onto $\mathrm{k}$-th plane, $\mathcal{N}$ normalizes $\mathbf{x}$'s range to [-1,1], and $\psi$ represents bilinear interpolations. Following \cite{kplanes}, different features from triplane are combined by Hadamard product as $F = \prod_k{F}_k$ to produce a final feature $F$. The aggregated features will be decoded into color $\mathbf{c}$ and density $\sigma$ by an MLP decoder $\MLP$. Thus $\MLP$ and $\mathcal{P}$ represent $\mathbf{\Theta}$ above in pose-NeRF optimization.
Besides photometric loss, a commonly used total variation loss $\mathcal{L}_{TV}$ \cite{plenoxels,kplanes} is applied, and the whole loss term $\mathcal{L}$ is formulated as:
% \vspace{-0.2cm}
\begin{equation}
\label{eqn:loss}
    \mathcal{L} = \mathcal{L}_{render}+\mathcal{L}_{TV}.
\end{equation}
\section{Method}

As mentioned above, the baseline that naively combines pose estimation with triplane has drawbacks on local errors in feature plane updating and inappropriate plane aggregation. To address that, we provide a novel pipeline as illustrated in \cref{fig:pipeline}. Our method can be divided into two stages. In the first stage, we input random triplane noise to the proposed triplane generator to generate different feature grids for scene representation while optimizing the camera poses (\cref{ms_conv}). Features interpolated from different planes are aggregated through the proposed DPA (\cref{distangled pa}). Colors and densities are derived from an MLP decoder. In the second stage, we adopt a warm-start training strategy for both efficient training and better scene representation (\cref{2stage}).

% \vspace{-0.2cm}
\subsection{Disentangled Triplane Generation Module}
\label{subsec:generator}

\begin{figure*}[t]
\centering
\includegraphics[width=\textwidth]{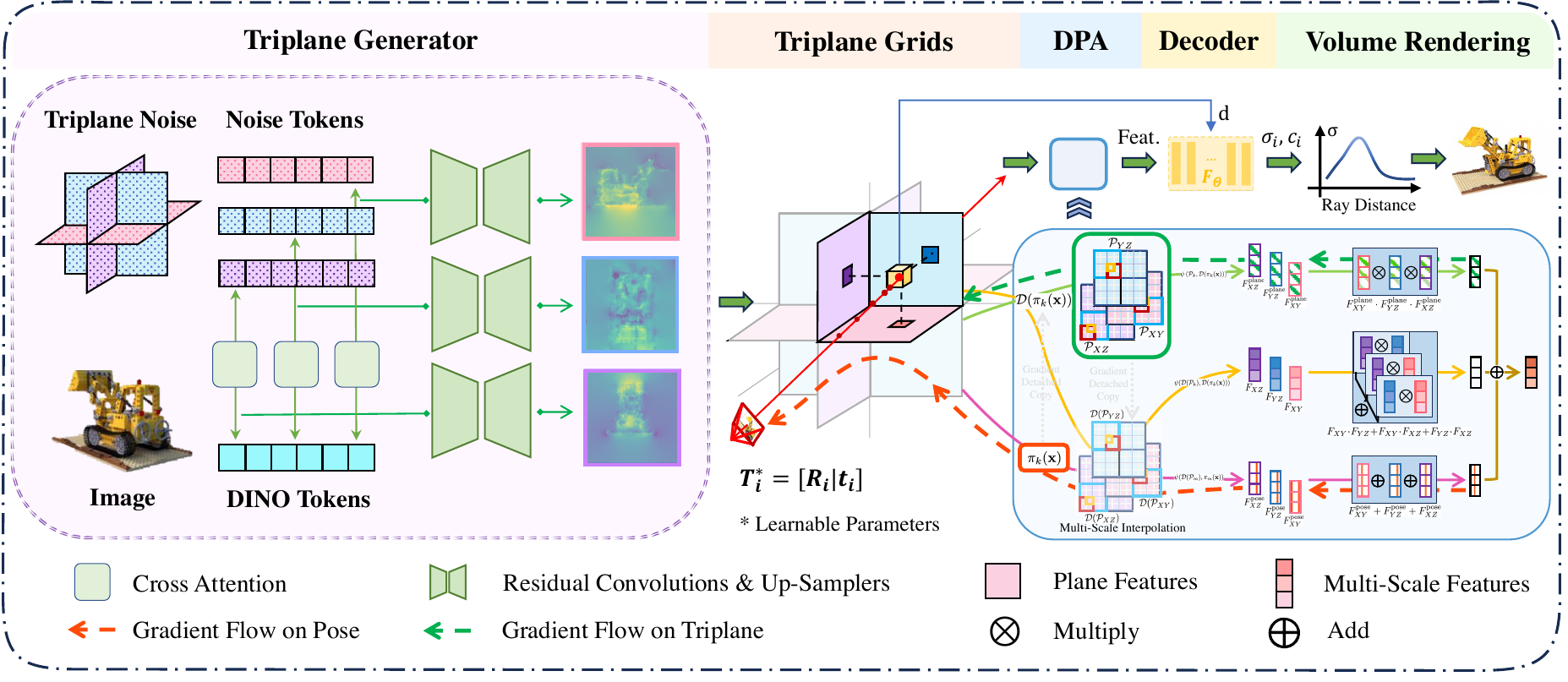}
% \vspace{-0.5cm}
\captionsetup{font=small}
\caption{\textbf{Overview pipeline of our proposed method.} In the 1st stage, we obtain triplane features from the triplane generator. In the 2nd stage, we discard the generator and transform it to direct updates on triplane feature grids.}
    \label{fig:pipeline}
    % \vspace{-0.6cm}
\end{figure*}
\textbf{Triplane Generator.}
\label{ms_conv}
As we mentioned in \cref{sec:intro}, the baseline joint estimation is prone to fall into local minima. To mitigate the local error, we propose a disentangled triplane generator for robust learning. 

As shown in \cref{fig:pipeline}, given a frozen triplane noise following a Gaussian distribution as input, we introduce a convolution-based neural network $\Gen$ as our triplane generator for extracting the plane features $\mathcal{P}$. Our motivation is to parameterize spatial feature grids as deep neural networks. Since the feature planes are highly decoupled for different perspectives of 3D scenes, features among different planes are anisotropic. Therefore, we apply three individual neural networks with the same structure but not shared parameters as disentangled generators to produce three different planes respectively.

A convolution-based network is applied as our triplane generator. The generator is randomly initialized for each scene and \textit{not} pre-trained, which still requires per-scene optimization. Detailed architecture of the generator can be found in the supplementary. As shown in \cref{fig:local}, compared to baseline, our triplane generator reduces errors from local updates and exhibits much less noise in the visualization of planar feature maps. In addition, the deep network prior will excavate the scene's geometric and texture features and embed them into triplane representation, instead of simply over-fitting the novel viewpoints.

It's worth noting that the input of triplane generator $\Gen$ is randomly initialized noise tokens $\token$ in order to introduce \textit{spatial priors}. 
During generation, the core features are maintained in a triplane structure from input to output, suiting the disentangled scene representation.

% \vspace{-0.3cm}
\subsubsection{Scene Texture Embedding Module.} 
% The Scene Texture Embedding (STE) module introduces 2D scene texture prior to the triplane generation, which enhances the texture representation in first stage triplane generation.
After the parameterization of the triplane with the generator, we present the Scene Texture Embedding module to enhance the triplane texture representation of the scene, thus mitigating the pose-NeRF ambiguous \cite{l2gnerf} problem. Inspired by \cite{LRM, PF-LRM}, we apply a DINOv2 \cite{dinov2} to encode the input image into patch-wise feature tokens. We found that without the need for several extracted feature maps for triplane generation, it is still capable of reducing the optimization ambiguity effectively. Therefore, we extract the feature $h_0$ of the first image $I_0$ for the scene texture prior.
% , which is utilized to bridge the 2D image feature with 3D representation. 

Next, a cross-attention is applied to incorporate the 2D feature into the triplane. During optimization, 2D feature tokens and 3D triplane tokens will autonomously learn the alignment between modalities, which allows for better integration. 
Specifically, we use the frozen triplane tokens as query, and the extracted feature tokens as key and value to obtain the attended triplane tokens $t_{in}$. Finally, $t_{in}$ is input to the triplane generator for the subsequent processing.

\subsection{Disentangled Plane Aggregation}

\label{distangled pa}
% remark
Features queried from different planes with a sampled 3D point $\textbf{x}$ will be aggregated before decoding and rendering. 
Here we analyze how the anisotropy in triplane influences pose in joint optimization, point out deficiencies in previous aggregation, and present our Disentangled Plane Aggregation as a solution.
\begin{figure}[t]
    \includegraphics[width=0.9\linewidth]{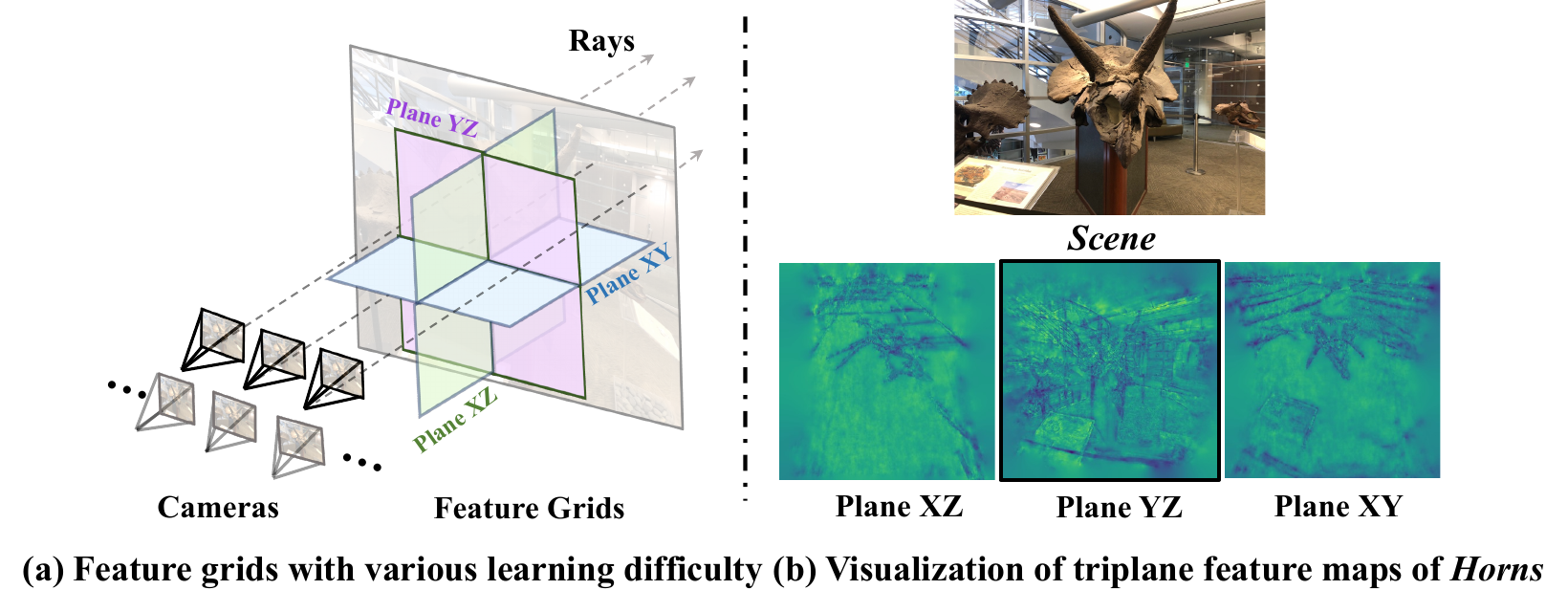}
    \centering
    % \vspace{-1mm}
    \caption{
    \textbf{Illustration of the training differences among feature grids.} (a) With this camera distribution, Plane YZ obtains more scene information and converges with less difficulty than the other planes. (b) We visualize the feature maps on the triplane of the same training steps. Compared to the other two planes, Plane YZ shows realistic texture to the original scene, demonstrating sufficient training.
    } 
    \label{fig:dpa}
    % \vspace{-0.7cm}
\end{figure}

Since the triplane representation is an interpretable explicit radiance field decomposing scenes into three orthogonal planes aligned with X, Y, and Z axes, each plane contains features of 3D scenes from orthogonal perspectives, which is influenced by different characteristics inherent to the scene.
Adequate captured images from varied viewpoints furnish comprehensive scene information to triplane across all perspectives, while the images from limited or unevenly distributed viewpoints provide disparate information to each plane.
% Taking forward-facing scenes for instance, 
For example, as depicted in \cref{fig:dpa}, angles of images or videos are generally consistent, with all objects primarily facing toward cameras. As scenes emphasize the front view of objects, Plane YZ in \cref{fig:dpa} gains richer scene information and is easier to learn.
During optimization, one plane that is easier to learn is more capable of fitting the scene and therefore provides better supervision for the pose. On the contrary, a plane that is more challenging to learn requires extra training iterations to accurately represent decoupled scene information, and therefore the probability of providing ambiguous supervision to the pose increases, resulting in pose optimizing collision during training. 

As mentioned in \cref{sec:preliminary} and \cref{subsec:generator}, given an 2D sample $\mathbf{u}=(u,v)$ on image space of camera $i$ and its homogeneous coordinates $\mathbf{\Bar{u}}=[\mathbf{u};1]^{\text{T}}$, we can get $\mathbf{x}=\mathcal{W}(z\mathbf{\Bar{u}},\mathbf{T}_i)$ at depth $z$ by space warping $\mathcal{W}$ and obtain features $F$ by interpolation and Hadamard product combination according to \cref{eqn:feat_k}. Here, we can rewrite \cref{eqn:render} as $\hat{I}(\mathbf{r})=\hat{I}(\MLP(F))$, and derive the partial derivative with respect to $\mathbf{T}_i$:
% \vspace{-0.2cm}
\begin{equation}
% \vspace{-0.3cm}
    \frac{\partial \hat{I_i}}{\partial \mathbf{T}_i}=\frac{\partial \hat{I_i}}{\partial \MLP} \frac{\partial \MLP}{\partial F} \frac{\partial F}{\partial \mathbf{x}} \frac{\partial \mathcal{W}(z\mathbf{\Bar{u}}, \mathbf{T}_i)}{\partial \mathbf{T}_i},
\end{equation}
which is formed via backpropagation and chain rules. Note that $\frac{\partial \hat{I_i}}{\partial \MLP}$ and $\frac{\partial \MLP}{\partial F}$ represent the rendering differentiation and weights of the MLP decoder relatively, $\frac{\partial \mathcal{W}(z\mathbf{\Bar{u}}, \mathbf{T}_i)}{\partial \mathbf{T}_i}$ on behalf of warping, we expand the remaining part $\frac{\partial F}{\partial \mathbf{x}}$ as:
% \vspace{-0.2cm}
\begin{equation}
\label{eqn:finalgrad}
    \frac{\partial F}{\partial \mathbf{x}}=\sum_{m}\left(\frac{\partial \Interp\left(\mathcal{P}_m, \pi_m(\mathbf{x})\right)}{\partial \mathbf{x}} \cdot \prod_{k\neq m}\Interp\left(\mathcal{P}_k, \pi_k(\mathbf{x})\right)\right),
    % \vspace{-0.2cm}
\end{equation}
\begin{wraptable}{r}{0.43\columnwidth}
% \vspace{-1.2cm}
    \scriptsize
\caption{\textbf{Ablation study on \textit{Horns} over different aggregations.}}
	\centering
\begin{tabular}{cccc}
\toprule
                       & Rot.(°) $\downarrow$ & Trans.($\times 100$)$\downarrow$ & PSNR$\uparrow$  \\ \midrule
\multicolumn{1}{c|}{Prod.}   & 1.002   & 0.348                & 21.64 \\
\multicolumn{1}{c|}{Sum}     & 0.794   & 0.172                & 21.29 \\
\multicolumn{1}{c|}{DPA}     & 0.296   & 0.102                & 23.66 \\ \bottomrule
\label{tab:dpa}
\end{tabular}
% \vspace{-1.2cm}
\end{wraptable}
where the subscript $m$ and $k$ are plane indexes, selected from set $\{XY,XZ,YZ\}$.
When updating poses, especially in early steps, gradients are influenced by the fluctuation on all feature planes, thus planes with inadequate supervision lead to ambiguity.
Another popular aggregation is sum, used in \cite{eg3d,hexplane}. However, as reported in \cref{tab:dpa}, while sum relieves pose optimizing collision, its ability to learn triplane to express 3D scene is inferior to product, which is consistent with \cite{kplanes}.

Aiming at both reducing collision to assist pose optimization and promoting the expressiveness of triplane, we design a disentangled aggregating strategy among planes. Dubbed as DPA, the expression of this aggregation is shown below:
\begin{equation}
\label{eqn:dpa}
\begin{split}
    \mathbf{DPA}(\mathcal{P},\mathbf{x})&=\prod_{k}\Interp\left(\mathcal{P}_k,\Detach(\pi_k(\mathbf{x}))\right)+\sum_{m}\Interp(\Detach(\mathcal{P}_m),\pi_m(\mathbf{x}))\\
    &+\sum_{k\neq m}^{k,m}(\Interp(\Detach(\mathcal{P}_k),\Detach(\pi_k(\mathbf{x}))) \cdot \Interp(\Detach(\mathcal{P}_m),\Detach(\pi_m(\mathbf{x}))))+1,
\end{split}
% \vspace{-0.3cm}
\end{equation}
where $\Detach(x)$ is the gradient-detached copy of $x$. Due to limited text space, we present detailed derivation of the above formula in supplemental materials.
Noticing that $\Detach(x)$ keeps the value, the output of DPA is equal to $\prod_{k}(F_k + 1)$ in a Hadamard product form, and thus preserves the expressiveness in triplane. In practice, we change the addition item $1$ to a hyperparameter $\lambda$ and so as the coefficients of items in \cref{eqn:dpa}. Allowing the triplane gradient obtained from product while the pose gradient from sum, DPA distributes information with disentanglement to pose and triplane, leading to more robust joint optimization.

% \vspace{-0.1cm}
\subsection{Two Stage Warm-Start Training}

\label{2stage}
After adopting the triplane generator, we find that once the camera poses are well initialized, the baseline joint pose-triplane estimation outperforms the training of adopting the generator from the beginning to the end, particularly in areas of high frequency. We argue that though the triplane generator mitigates local updating errors, it suffers from implicit constraints in different patches on the plane grids caused by the generator, thus introducing excessive smoothing on the feature grids. Therefore, we propose a warm-start two-stage training strategy to switch to direct plane optimization for better NVS quality. Instead of initializing the parameters from scratch in the second stage, we leverage the learned parameters in the first stage. The MLP decoder and camera poses are inherited directly as they work for the same triplane which contains no gaps between different stages. For the 2nd-stage triplane, we discard the generator and perform a forward inference to obtain the final 1st-stage triplane. Following \cite{kplanes, instantngp}, we adopt multiple feature grids from different resolutions in a coarse-to-fine manner, thereby enhancing the structural and detailed representation. The 2nd-stage triplane with multiple scales will be obtained by bilinear interpolation from the 1st-stage triplane, where direct triplane optimization will be conducted in the second stage.

\begin{figure}[!tb]
  \centering
  \includegraphics[width=\textwidth]{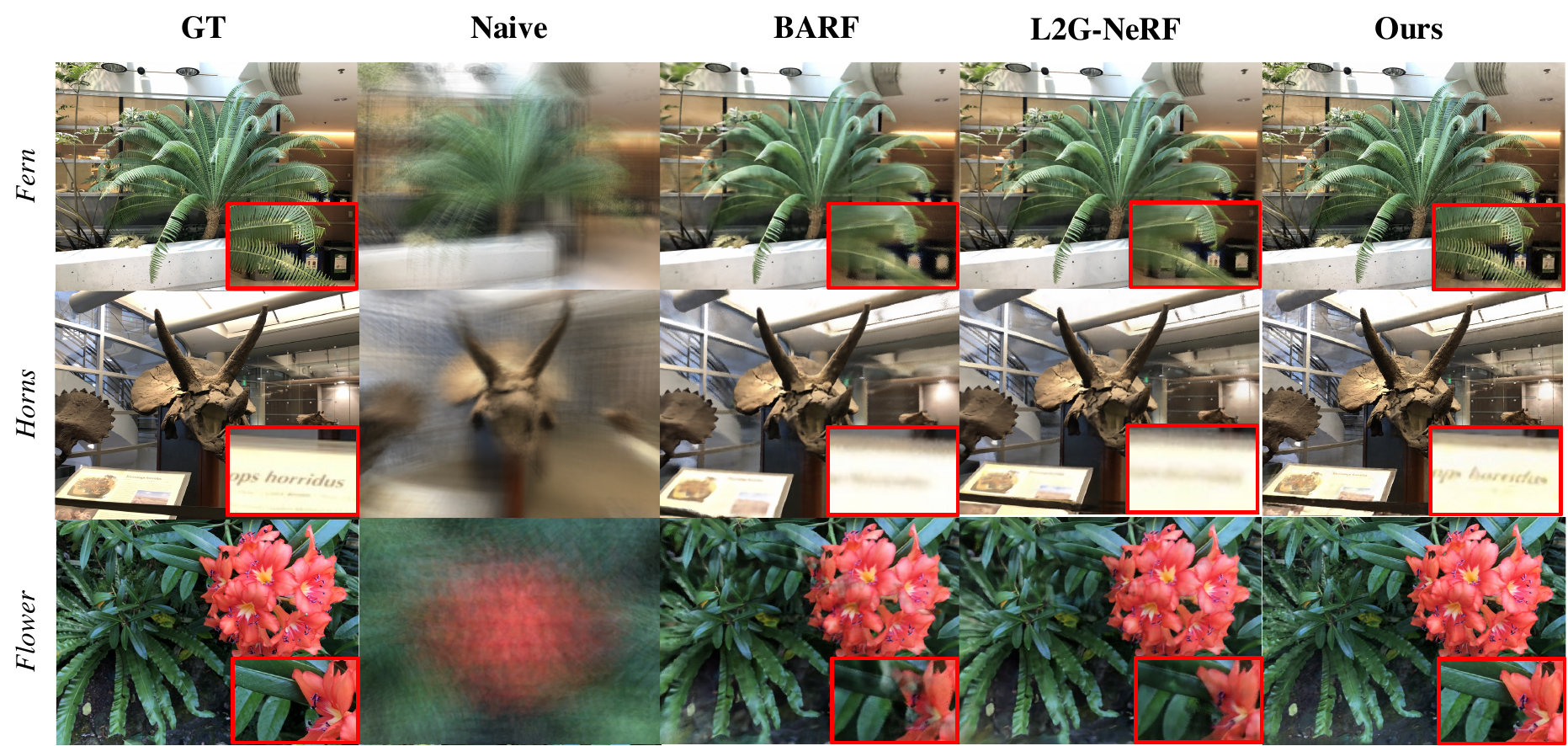}
  % \vspace{-0.4cm}
  \caption{
    \textbf{Qualitative results of novel view synthesis on LLFF dataset.} Naive represents reference K-Plane \cite{kplanes} trained with unknown camera poses.
  }
  \label{fig:res_llff}
  % \vspace{-0.3cm}
\end{figure}
% \vspace{-0.2cm}
\section{Experiments}
% \vspace{-0.2cm}
\label{sec:exp}
In this section, we present the performance and analysis of our proposed method on two public datasets: LLFF\cite{llff} and NeRF-Synthetic dataset\cite{mildenhall2021nerf}. Next, we compare our results on both pose accuracy and novel view synthesis quality with other joint optimizing works containing BARF\cite{lin2021barf}, GARF\cite{chng2022garf}, L2G-NeRF\cite{l2gnerf} and HASH-BARF\cite{robust_hash} in \cref{sec: llff} and \cref{sec: blender}. 
% Since the implementation of GARF and HASH-BARF are unavailable, we directly use the results reported in their paper for comparison. 
Lastly, we conduct ablation studies to validate the effectiveness of our proposed modules in \cref{sec: ablation}.
% , and further compare our model with the reference COLMAP-based triplane-NeRF \cite{kplanes}. 
% More implementation details can be found in the supplementary.

% \vspace{-0.4cm}
\subsection{Results on LLFF dataset}
\label{sec: llff}

\subsubsection{Experimental Settings.} For LLFF dataset \cite{llff}, the resolution of images is set to 480 $\times$ 640 \cite{lin2021barf,robust_hash} and the train/test splits are defined as \cite{lin2021barf,l2gnerf,chng2022garf}. We initialize the translation of each camera as a zero vector and the rotation matrix as an identity matrix. Our model is trained for 70k steps and switched to the second stage at step 4000. In the testing phase, our model performs the same test-time photometric pose optimization \cite{lin2021barf,l2gnerf} before evaluating view synthesis quality. The ground-truth poses are estimated from COLMAP\cite{colmap}. We further compare our method to the reference baseline K-Planes \cite{kplanes} without camera poses. All methods are trained and evaluated under the same settings.

\begin{table}[tb]
  \caption{\textbf{Quantitative results of novel view synthesis on LLFF dataset.}}

  \label{tab:llff_psnr}
  \centering
  % \vspace{-0.3cm}
  \scriptsize
  \begin{adjustbox}{width=1\textwidth} %{width=1\textwidth}
  \begin{tabular}{ c  c c c c c  c c c c c c c c }
    \toprule
    
    \multirow{4}{*}{\textbf{Scene}} 
    % & \multicolumn{8}{c}{Camera Pose Registration} 
    & \multicolumn{10}{c}{\textbf{Novel View Synthesis Quality}} \\ 
    
    \cmidrule(r){2-11} 
    & \multicolumn{5}{c}{PSNR $\uparrow$}
    & \multicolumn{5}{c}{SSIM $\uparrow$}\\
    \cmidrule(r){2-6}
    \cmidrule(r){7-11}
    & BARF\cite{lin2021barf}
    & GARF\cite{chng2022garf}
    & L2G\cite{l2gnerf}
    & HASH\cite{robust_hash}
    & Ours
    
    & BARF\cite{lin2021barf}
    & GARF\cite{chng2022garf}
    & L2G\cite{l2gnerf}
    & HASH\cite{robust_hash}
    & Ours
    
    \\
    \midrule 
    
    Fern 
    
    & 23.79
    & 24.51
    & 24.57
    & 24.62
    & \textbf{25.70}
    
    & 0.710
    & 0.740
    & 0.750
    & 0.743
    & \textbf{0.837}
    
    \\
    
    Flower 
    
    & 23.37
    & 26.40
    & 24.90
    & 25.19
    & \textbf{27.06}
    
    & 0.698
    & 0.790
    & 0.740
    & 0.744
    & \textbf{0.830}
    
    \\
    
    Fortress 
    
    & 29.08
    & 29.09
    & 29.27
    & 30.14
    & \textbf{30.79}
    
    & 0.823
    & 0.820
    & 0.840
    & 0.901
    & \textbf{0.905}
    
    \\
    
    Horns 
    
    & 22.78
    & 22.54
    & 23.12
    & 22.97
    & \textbf{23.66}
    
    & 0.727
    & 0.690
    & 0.740
    & 0.736
    & \textbf{0.828}
    \\
    
    Leaves 
    
    & 18.78
    & 19.72
    & 19.02
    & 19.45
    & \textbf{20.43}
    
    & 0.537
    & 0.610
    & 0.560
    & 0.607
    & \textbf{0.708}
    
    \\
    
    Orchids 
    
    & 19.45
    & 19.37
    & 19.71
    & 20.02
    & \textbf{20.24}
    
    & 0.574
    & 0.570
    & 0.610
    & 0.610
    & \textbf{0.660}
    
    \\
    
    Room 
    
    & 31.95
    & 31.90
    & 32.25
    & 32.73
    & \textbf{33.95}
    
    & {0.949}
    & 0.940
    & 0.950
    & 0.968
    & \textbf{0.970}
    
    \\
    
    T-Rex 
    
    & 22.55
    & 22.86
    & 23.49
    & 23.19
    & \textbf{25.35}
    
    & 0.767
    & 0.800
    & 0.800
    & 0.866
    & \textbf{0.889}
    
    \\
    
    \midrule
    Mean 
    
    & 23.97
    & 24.55
    & 24.54
    & 24.79
    & \textbf{25.90}
    
    & 0.723
    & 0.745
    & 0.750
    & 0.772
    & \textbf{0.828}
    
    \\
    
    \bottomrule
  \end{tabular}
  \end{adjustbox}
  % \vspace{-0.3cm}
\end{table}

\begin{table}[t]
  \caption{\textbf{Quantitative results of camera pose estimation on LLFF dataset.}}
  \label{tab:llff_pose}
  \centering
  \setlength{\tabcolsep}{4pt}
  % \vspace{-0.3cm}
  \scriptsize
  \begin{adjustbox}{width=1\textwidth}
  \begin{tabular}{ c  c c c c c  c c c c c  }
    \toprule
    
    \multirow{4}{*}{\textbf{Scene}} 
    & \multicolumn{10}{c}{\textbf{Camera Pose Estimation}} \\ 
     
    \cmidrule(r){2-11} 
    & \multicolumn{5}{c}{Rotation $({}^{\circ})$ $\downarrow$}
    & \multicolumn{5}{c}{Translation $(\times 100)$ $\downarrow$}\\
    \cmidrule(r){2-6}
    \cmidrule(r){7-11}
    & BARF\cite{lin2021barf}
    & GARF\cite{chng2022garf}
    & L2G\cite{l2gnerf}
    & HASH\cite{robust_hash}
    & Ours
    
    & BARF\cite{lin2021barf}
    & GARF\cite{chng2022garf}
    & L2G\cite{l2gnerf}
    & HASH\cite{robust_hash}
    & Ours
    
    \\
    \midrule 
    
    Fern 
    
    & 0.191
    & 0.470
    & 0.200
    & \textbf{0.110}
    & 0.243
    
    & \textbf{0.102}
    & 0.250
    & 0.180
    & \textbf{0.102}
    & 0.166
    
    \\
    
    Flower 
    
    & 0.251 % 0.47?
    & 0.460
    & 0.330
    & 0.301
    & \textbf{0.139}
    
    & 0.224
    & 0.220
    & 0.240
    & 0.211
    & \textbf{0.179}
    
    \\
    
    Fortress 
    
    & 0.479 % 0.17?
    & \textbf{0.030}
    & 0.250
    & 0.211
    & 0.612
    
    & 0.364
    & 0.270
    & 0.250
    & \textbf{0.241}
    & 0.385
    
    \\
    
    Horns 
    
    & 0.304 % 3.50? 
    & \textbf{0.030}
    & 0.220
    & 0.049
    & 0.296
     
    & 0.222 % 1.32?
    & 0.210
    & 0.270
    & 0.209
    & \textbf{0.102}
    \\
    
    Leaves 
    
    & 1.272
    & \textbf{0.130}
    & 0.790
    & 0.840
    & 0.329
    
    & 0.249
    & 0.230
    & 0.340
    & \textbf{0.228}
    & 0.239
    
    \\
    
    Orchids 
    
    & 0.627
    & 0.430
    & 0.670
    & 0.399
    & \textbf{0.226}
    
    & 0.404
    & 0.410
    & 0.410
    & 0.386
    & \textbf{0.235}
    
    \\
    
    Room 
    
    & 0.320
    & 0.420
    & 0.300
    & 0.271
    & \textbf{0.184}
    
    & 0.270
    & 0.320
    & 0.230
    & 0.213
    & \textbf{0.095}
    
    \\
    
    T-Rex 
    
    & 1.138 % 0.66?
    & 0.660
    & 0.890
    & 0.894
    & \textbf{0.039}
    
    & 0.720 % 0.36?
    & 0.480
    & 0.640
    & 0.474
    & \textbf{0.149}
    
    \\
    
    \midrule
    Mean 
    
    & 0.573
    & 0.329
    & 0.460
    & 0.384
    & \textbf{0.259}
    
    & 0.331
    & 0.299
    & 0.320
    & 0.258
    & \textbf{0.194}
    
    \\
    
    \bottomrule
  \end{tabular}
  % \vspace{-1.0cm}
  \end{adjustbox}
\end{table}
% \vspace{-0.2cm}
\subsubsection{Results.} Quantitative results of novel view synthesis and pose estimation are shown in \cref{tab:llff_psnr} and \cref{tab:llff_pose}. The PSNR and SSIM are reported for novel view synthesis quality, and the rotation error (in degree) and translation error (scaled by 100) are reported for pose estimation accuracy. Qualitative results are shown in \cref{fig:res_llff} for visual comparison. Without pose prior and optimization (Naive), it is impossible to reconstruct the scene with common triplane radiance fields.

% \usepackage{multirow}
% \vspace{-0.2cm}
\subsection{Results on NeRF-Synthetic dataset}
\label{sec: blender}

\subsubsection{Experimental Settings.} In our implementation, the rendered images are resized to 400 × 400 resolution, and the train/test splits are the same as \cite{lin2021barf,l2gnerf,robust_hash,chng2022garf}. We follow \cite{lin2021barf} to impose additive Gaussian noise $\xi \in \mathfrak{s e}(3)$ and $\xi \sim \mathcal{N}(\mathbf{0}, 0.15 \mathbf{I})$ to the ground truth poses as initialization. Our model is trained for 60k steps and switched to the second stage at 2000 steps. The settings of test-time pose optimization are the same as the LLFF dataset. Note that L2G-NeRF \cite{l2gnerf} is implemented with a different way of the pose noise perturbation, we re-implement L2G-NeRF's noise perturbation according to BARF to ensure the fairness of comparisons. We further show the results of the reference baseline K-Planes with the same noisy pose initialization.

% \vspace{-0.4cm}
\subsubsection{Results.} 
\begin{figure}[tb]
  \centering
  \includegraphics[width=\textwidth]{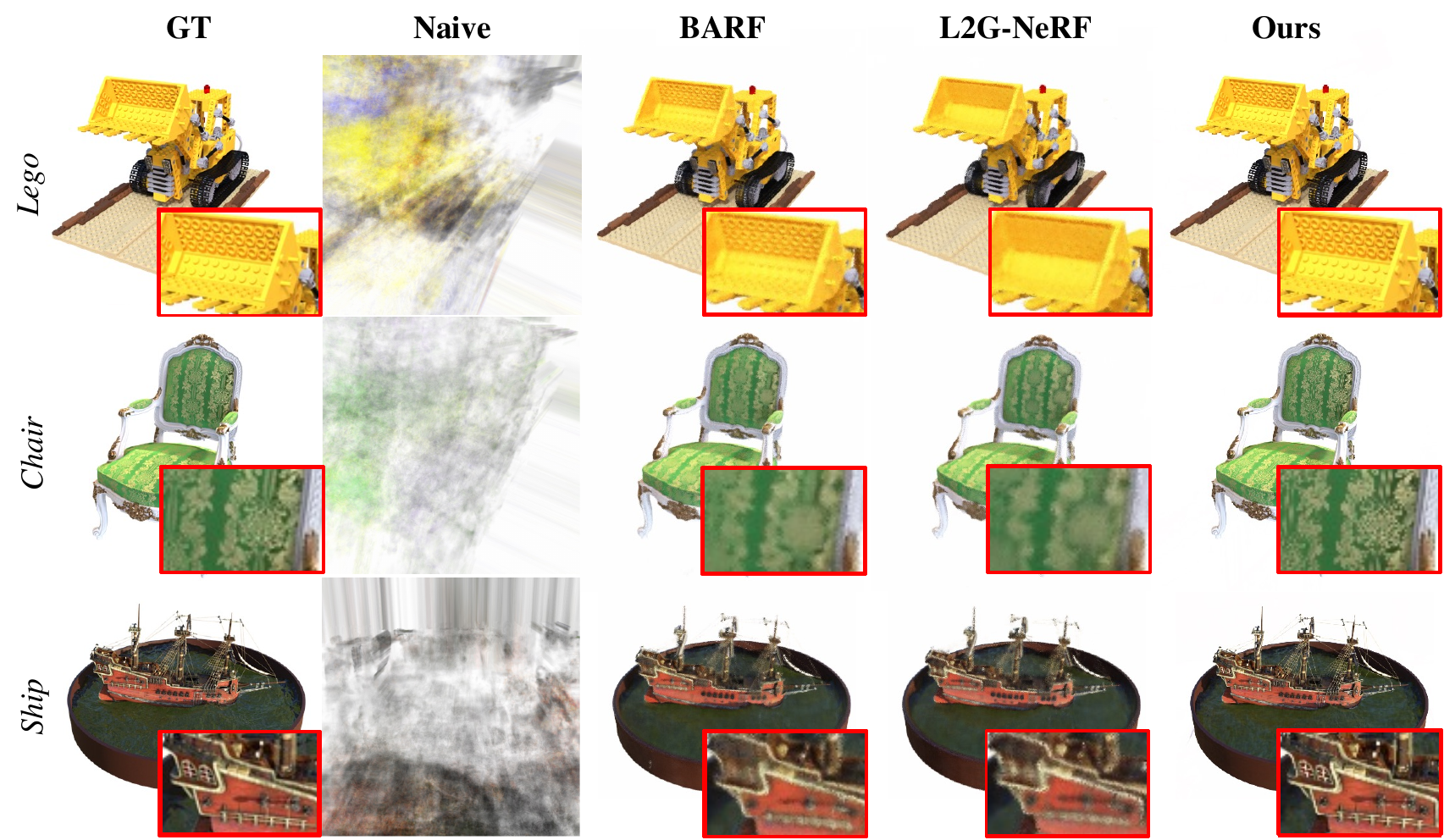}
  % \vspace{-0.2cm}
  \caption{
    \textbf{Qualitative results of novel view synthesis on NeRF-synthetic dataset.} Naive represents reference baseline K-Plane \cite{kplanes} is trained with noisy camera poses.
  }
  \label{fig:res_blender}
  % \vspace{-0.4cm}
\end{figure}

The quantitative results of NVS and pose estimation are presented in \cref{tab:blender_psnr} and \cref{tab:blender_pose}, where the same metrics are reported as in the LLFF dataset. The qualitative NVS results are shown in \cref{fig:res_blender}. The original triplane method K-Planes fails to perform reliable scene reconstruction without accurate camera poses, while our proposed method still performs robust novel view synthesis.

% \vspace{-0.2cm}
\subsection{Ablation Study}
\label{sec: ablation}
In this section, we analyze the effectiveness of our proposed core components. The ablation studies are shown in \cref{tabs:ablation}. 
% \vspace{-0.2cm}
\subsubsection{Effectiveness of Triplane Generator.} 
The rows (e) and (c) show the results of the baseline and that with the triplane generator. The baseline method suffers from severe local minima due to incorrect local updating. After introducing the triplane generator, the quality of joint estimation substantially improved. 

\subsubsection{Effectiveness of Disentangled Plane Aggregation.}
After introducing DPA to the baseline, as shown in (e) and (d), indicating that the pose estimation is subject to the conflict between different planes. Therefore the proposed DPA can improve the accuracy of pose estimation and thus bring better NVS quality. Meanwhile, after removing DPA from (b) to (c), there is a certain decrease in the joint optimization results.

\subsubsection{Effectiveness of Two-Stage Warm-Start Training.}
Without two-stage training, over-smoothness caused by the triplane generator degrades the image rendering quality. In addition, as shown in \cref{table:traintime}, we conduct runtime comparisons to show the trade-off between efficiency and performance. Our method incurs a larger time consumption when using only the first stage of training while the quality of viewpoint rendering is degraded. Our full model improves the image rendering quality with comparable time consumption to baseline direct 
optimization.

\begin{table*}[tb]
  \caption{\textbf{Quantitative results of novel view synthesis on NeRF-Synthetic.}}
  % \vspace{-0.3cm}
  % \setlength{\tabcolsep}{4pt}
  \scriptsize
  \label{tab:blender_psnr}
  \centering
  \begin{adjustbox}{width=1\textwidth}
  \begin{tabular}{ c  c c c c c  c c c c c c c c }
    \toprule
    
    \multirow{4}{*}{Scene} 
    & \multicolumn{10}{c}{Novel View Synthesis} \\ 
    
    \cmidrule(r){2-11} 
    & \multicolumn{5}{c}{PSNR $\uparrow$}
    & \multicolumn{5}{c}{SSIM $\uparrow$}\\
    \cmidrule(r){2-6}
    \cmidrule(r){7-11}
    & BARF\cite{lin2021barf}
    & GARF\cite{chng2022garf}
    & L2G\cite{l2gnerf}
    & HASH\cite{robust_hash}
    & Ours
    
    & BARF\cite{lin2021barf}
    & GARF\cite{chng2022garf}
    & L2G\cite{l2gnerf}
    & HASH\cite{robust_hash}
    & Ours
    
    \\
    \midrule 
    
    Chair
    & 31.16
    & 31.32
    & 34.28
    & 31.95
    & \textbf{37.78}
    
    & 0.954
    & 0.959
    & 0.982
    & 0.962
    & \textbf{0.992}
    \\
    
    Drum
    
    & 23.91
    & 24.15
    & 25.39
    & 24.16
    & \textbf{26.10}
    
    & 0.900
    & 0.909
    & 0.901
    & 0.912
    & \textbf{0.926}
    \\
    
    Ficus 
    & 26.26
    & 26.29
    & 27.45
    & 28.31
    & \textbf{33.12}
    
    & 0.934
    & 0.935
    & 0.940
    & 0.943
    & \textbf{0.979}
    \\
    
    Hotdog 
    & 34.54
    & 34.69
    & 34.03
    & 35.41
    & \textbf{35.98}
    
    & 0.970
    & 0.972
    & 0.967
    & \textbf{0.981}
    & 0.976
    \\
    
    Lego 
    & 28.33
    & 29.29
    & 27.66
    & 31.65
    & \textbf{32.19}
    
    & 0.927
    & 0.925
    & 0.922
    & 0.973
    & \textbf{0.976}
    \\
    
    Materials 
    & 27.84
    & \textbf{27.91}
    & 26.01
    & 27.14
    & 25.78
    
    & 0.936
    & \textbf{0.941} 
    & 0.920
    & 0.911
    & 0.920
    \\
    
    Mic 
    & 31.18
    & 31.39
    & 32.61
    & 32.33
    & \textbf{33.88}
    
    & 0.969
    & 0.971
    & 0.971
    & 0.975
    & \textbf{0.987}
    \\
    
    Ship 
    & 27.50
    & 27.64
    & 28.30
    & 27.92
    & \textbf{29.80}
    
    & 0.849
    & 0.862
    & 0.789
    & 0.879
    & \textbf{0.903}
    \\
    
    \midrule
    Mean 
    & 28.84
    & 28.96
    & 29.47
    & 29.86
    & \textbf{31.83}
    
    & 0.930
    & 0.935
    & 0.924
    & 0.943
    & \textbf{0.957}
    
    \\
    
    \bottomrule
  \end{tabular}
  \end{adjustbox}
  % \vspace{-0.3cm}
\end{table*}
\begin{table*}[t]
  \caption{\textbf{Quantitative results of pose estimation on NeRF-Synthetic.}}
  \vspace{-0.3cm}
  \label{tab:blender_pose}
  \centering
  \scriptsize
  \begin{adjustbox}{width=1\textwidth}
  % \vspace{-0.3cm}
  \begin{tabular}{ c  c c c c c  c c c c c c c c }
    \toprule
    
    \multirow{4}{*}{Scene} 
    & \multicolumn{10}{c}{Camera Pose Estimation} \\
    
    \cmidrule(r){2-11} 
    & \multicolumn{5}{c}{Rotation $({}^{\circ})$ $\downarrow$}
    & \multicolumn{5}{c}{Translation ($\times 100$) $\downarrow$}\\
    \cmidrule(r){2-6}
    \cmidrule(r){7-11}
    & BARF\cite{lin2021barf}
    & GARF\cite{chng2022garf}
    & L2G\cite{l2gnerf}
    & HASH\cite{robust_hash}
    & Ours
    
    & BARF\cite{lin2021barf}
    & GARF\cite{chng2022garf}
    & L2G\cite{l2gnerf}
    & HASH\cite{robust_hash}
    & Ours
    
    \\
    \midrule 
    
    Chair 
    & 0.096
    & 0.113
    & 0.118
    & 0.085
    & \textbf{0.036}
    
    & 0.428
    & 0.549
    & 0.495
    & 0.365
    & \textbf{0.186}
    
    \\
    
    Drum 
    & 0.043
    & 0.052
    & 0.070
    & 0.041
    & \textbf{0.032}
    
    & 0.225
    & 0.232
    & 0.340
    & 0.214
    & \textbf{0.197}
    
    \\
    
    Ficus 
    & 0.085
    & 0.081
    & 0.168
    &  0.079
    & \textbf{0.058}
    
    & 0.474
    & 0.461
    & 1.037
    & 0.479
    & \textbf{0.299}
    
    \\
    
    Hotdog 
    & 0.248
    & 0.235
    & 0.661
    & 0.229
    & \textbf{0.226}
    
    & 1.308
    & 1.123
    & 4.283
    & 1.123
    & \textbf{0.976}
    \\
    
    Lego 
    & 0.082
    & 0.101 
    & 0.088
    & 0.071
    & \textbf{0.033}
    
    & 0.291
    & 0.299
    & 0.399
    & 0.272
    & \textbf{0.141}
    
    \\
    
    Materials 
    & 0.844
    & 0.842
    & 0.805
    & 0.852
    & \textbf{0.486}
    
    & {2.692}
    & 2.688
    & 2.510
    & 2.743
    & \textbf{1.537}
    
    \\
    
    Mic 
    & 0.071
    & 0.070
    & 0.080
    & \textbf{0.068}
    & 0.124
    
    & 0.301
    & 0.293
    & 0.331
    & \textbf{0.287}
    & 1.554
    
    \\
    
    Ship 
    & {0.075}
    & \textbf{0.073}
    & 0.163
    & 0.079
    & 0.255
    
    & 0.326
    & 0.310
    & 0.585
    & \textbf{0.287}
    & 0.340
    
    \\
    
    \midrule
    Mean 
    & 0.193
    & 0.195
    & 0.269
    & {0.189}
    & \textbf{0.156}
    
    & 0.756
    & 0.744
    & 1.248
    & {0.722}
    & \textbf{0.654}
    
    \\
    
    \bottomrule
  \end{tabular}
  % \vspace{-0.2cm}
  \end{adjustbox}
\end{table*}

\begin{table}[]
  \centering
  % \label{tab:kp_naive}
  \setlength{\tabcolsep}{5pt}
  \caption{Comparison of the reference K-Planes\cite{kplanes} (Ref.), naive joint pose-triplane optimization (Base), and our proposed method (Ours). \# represents the reference K-Planes is trained with ground-truth camera poses.} 
  % \vspace{-0.2cm}
  % \footnotesize
  \begin{adjustbox}{width=0.95\textwidth}
\begin{tabular}{c c c c c c c   c c c c}
    \hline
  \label{tab:kp_naive}
\multirow{2}{*}{Scenes}       & \multicolumn{3}{c}{PSNR$\uparrow$} & \multicolumn{3}{c}{SSIM$\uparrow$} & \multicolumn{2}{c}{Rot.(°)$\downarrow$} & \multicolumn{2}{c}{Trans.($\times 100) \downarrow$} \\     \cmidrule(r){2-7}
    \cmidrule(r){8-11}
                              & Ref.\#     & Base      & Ours      & Ref.\#      & Base      & Ours      & Base               & Ours               & Base                     & Ours                     \\  \cmidrule(r){1-7}
    \cmidrule(r){8-11}
\multicolumn{1}{c}{Fern}     & 24.44      & 23.94     & 25.70     & 0.828       & 0.782     & 0.837     & 1.852              & 0.243              & 0.428                    & 0.166                    \\
\multicolumn{1}{c}{Flower}   & 27.42      & 24.88     & 27.06     & 0.872       & 0.764     & 0.830     & 8.624              & 0.139              & 0.733                    & 0.179                    \\
\multicolumn{1}{c}{Fortress} & 29.36      & 28.09     & 30.79     & 0.804       & 0.836     & 0.905     & 2.247              & 0.612              & 1.660                    & 0.385                    \\
\multicolumn{1}{c}{Horns}    & 28.64      & 20.82     & 23.66     & 0.892       & 0.543     & 0.828     & 63.92              & 0.296              & 22.93                    & 0.102                    \\
\multicolumn{1}{c}{Leaves}   & 20.22      & 16.46     & 20.43     & 0.746       & 0.427     & 0.708     & 173.6              & 0.329              & 6.799                    & 0.239                    \\
\multicolumn{1}{c}{Orchids}  & 19.58      & 14.03     & 20.22     & 0.676       & 0.241     & 0.660     & 17.48              & 0.226              & 5.283                    & 0.235                    \\
\multicolumn{1}{c}{Room}     & 34.07      & 21.68     & 33.95     & 0.957       & 0.716     & 0.970     & 174.2              & 0.184              & 6.139                    & 0.095                    \\
 \multicolumn{1}{c}{T-Rex}    & 24.14      & 21.86     & 25.35     & 0.915       & 0.675     & 0.889     & 84.62              & 0.039              & 21.26                    & 0.149                    \\  \cmidrule(r){1-7}
    \cmidrule(r){8-11}
\multicolumn{1}{c}{Mean}     & \textbf{25.98}      & 21.47     & 25.90     & \textbf{0.847}       & 0.623     & 0.828     & 65.82              & \textbf{0.259}              & 8.154                    & \textbf{0.194}                   \\ \hline
\end{tabular}
\end{adjustbox}

\end{table}

\begin{table*}[!t]

\setlength{\tabcolsep}{4pt}

  \begin{minipage}{0.46\columnwidth}
  % \captionsetup{font=scriptsize}
  % \vspace{-0.1cm}
    \caption{Ablation study of the proposed components on LLFF dataset\cite{llff}.}

    \centering
    % \hrule
    % \scriptsize
    \resizebox{0.965\textwidth}{!}{%
  \begin{tabular}{ c  c c c | c c c  }
    \toprule
    \multirow{2}{*}{}
    & $\ $ TriPlane $\ $
    & $\ $ Disentangled $\ $
    & $\ $ Two- $\ $
    % \hspace{1pt} 
    & {Rot.} 
    & {Trans.} 
    & {PSNR} 
    \\  
    & $\ $ Generator $\ $
    & $\ $ Plane Agg. $\ $
    & $\ $ Stage $\ $
    % \hspace{1pt} 
    & {$\downarrow$} 
    & {$\downarrow$} 
    & {$\uparrow$} 
    \\  
    \midrule
     
    (a)
    & \checkmark & \checkmark & \checkmark
    & \textbf{0.259}
    & \textbf{0.194}
    & \textbf{25.90}
    \\ 
    
    (b)
    &  \checkmark & \checkmark & 
    & 0.634
    % & 0.371
    & 0.328
    & 25.21
    \\
     
    (c)
    & \checkmark  &  & 
    & 0.660
    & 0.356
    & 24.93
    \\

    (d) 
    &   & \checkmark &
    & 6.180
    & 2.708
    & 23.41
    \\
    
    (e) 
    &   &  &
    & 65.82
    & 8.154
    & 21.47
    \\
    
    \bottomrule
  \end{tabular}
    }
    \label{tabs:ablation}
  \end{minipage}\hfill % maximize the horizontal separation
    \begin{minipage}{0.5\columnwidth}
    \centering
    % \hrule
    % \captionsetup{font=scriptsize}
    % \vspace{-0.1cm}
    \caption{Runtime comparisons on \textit{Drums}. Training Times are in the format of hh: mm.}
    % \vspace{-0.2cm}
    \resizebox{\textwidth}{!}{%
    \scriptsize 
\begin{tabular}{cccc}

\hline
 Methods                & Iters. & Training Time & PSNR$\uparrow$ \\ \hline
BARF\cite{lin2021barf}             & 200k   & 11:49              & 23.91          \\
HASH-BARF\cite{robust_hash}        & 200k   & \textbf{00:36}              & 24.16          \\ \hline
Ours-w/o $1^{st}$ stage & 60k    & 00:59              & 19.93          \\
Ours-w/o $2^{nd}$ stage & 60k    & 04:00              & 24.98          \\
Ours     & \textbf{60k}    & 01:21              & \textbf{26.10}          \\ \hline
\end{tabular}
    }
    \label{table:traintime} 
  \end{minipage}\vspace{-0.2cm}

\end{table*}

% \vspace{-0.3cm}
\subsubsection{Comparisons with Baseline Joint Estimation.} As shown in \cref{tab:kp_naive}, we conduct experiments to compare our full model, reference K-Planes \cite{kplanes} with COLMAP poses, and direct joint pose-triplane optimization baseline to illustrate the effectiveness. Note that the train/test split of the original K-Planes implementation is different from ours, we re-implement the evaluation as \cite{lin2021barf,l2gnerf,chng2022garf} for fair comparison. We parameterize the camera poses and integrate them into K-Planes as our baseline implementation. Our approach exhibits significant improvements in pose estimation and novel view rendering and even surpasses K-Planes with COLMAP poses, which further demonstrates the effectiveness of our method.

\section{Conclusion}

In this paper, we propose a novel algorithm that allows joint estimation of camera pose and disentangled scene representation to reduce the computational cost of neural renderings. We first propose the Disentangled Triplane Generation Module to parameterize the triplane with a convolution-based generator to mitigate local updating errors in the direct joint optimization baseline. Then we analyze the effect of triplane feature aggregation on camera poses and propose Disentangled Plane Aggregation to reduce the entanglement between feature planes and camera poses. Finally, we adopt a two-stage warm-start strategy to tackle the oversmoothing on the feature planes caused by the generator. Comprehensive evaluations demonstrate that our proposed method achieves state-of-the-art performance and rapid convergence with noisy or unknown poses.
% \vspace{-0.2cm}
% \subsubsection{Limitations and Future Works.}Although our method can recover the camera pose and triplane radiance fields effectively, there is still room for improvement. In the future, we will explore introducing more powerful 3D representations such as 3D Gaussian Splatting \cite{3dgs} into the joint optimization pipeline and try to further reduce the dependence on camera pose initialization.

\section*{Acknowledgments}

This work is financially supported by Outstanding Talents Training Fund in Shenzhen, Shenzhen Science and Technology Program-Shenzhen Cultivation of Excellent Scientific and Technological Innovation Talents project (Grant No. RCJC20200714114435057), Shenzhen Science and Technology Program-Shenzhen Hong Kong joint funding project (Grant No. SGDX20211123144400001), Guangdong Provincial Key Laboratory of Ultra High Definition Immersive Media Technology, National Natural Science Foundation of China U21B2012, R24115SG MIGU-PKU META VISION TECHNOLOGY INNOVATION LAB.
Jianbo Jiao is supported by the Royal Society Short Industry Fellowship (SIF$\setminus$R1$\setminus$231009).
% \bibliography{main}

\newpage

\renewcommand\thesection{\Alph{section}}

\title{Supplementary Material for: \\
Disentangled Generation and Aggregation \\ 
for Robust Radiance Fields}

% \titlerunning{Supplementary Material}
\titlerunning{Disentangled Generation and Aggregation
for Robust Radiance Fields}

\author{Shihe Shen \inst{1}\textsuperscript{$\star$} \and
Huachen Gao\inst{1}\textsuperscript{$\star$} \and Wangze Xu \inst{1} \and Rui Peng \inst{1,2} \and 
\\ Luyang Tang \inst{1,2} \and Kaiqiang Xiong \inst{1,2} \and Jianbo Jiao \inst{3} \and Ronggang Wang\inst{1,2,}\textsuperscript{\Letter} }

\authorrunning{S. Shen, H. Gao et al.}

\institute{School of Electronic and Computer Engineering, Peking University \and Peng Cheng Laboratory \and School of Computer Science, University of Birmingham\\
\email{\{shshen0308, gaohuachen712\}@gmail.com rgwang@pkusz.edu.cn}
\blfootnote{\textsuperscript{$\star$} Equal contribution.}
}

\maketitle

\section{Detailed Architecture of Triplane Generator}

As we mentioned in the previous chapter, the shape of triplane noise tokens is set to ($3 \times 8 \times 20\times 20$), where 3 represents the number of feature planes, 8 represents the hidden dimension and $20 \times 20$ is the spatial shape. The triplane tokens are reshaped to ($3 \times 400 \times 8$), and 
each plane ($1 \times 400 \times 8$) is applied as the query to do cross-attention with the extracted DINO feature tokens ($1 \times 3889 \times 384$) separately. The attended tokens are then reshaped back to ($3 \times 8 \times 20\times 20$) for the input of the triplane generator. The generator for each plane is composed of one mid-block and $L$ up-sample-blocks ($L=5$). The mid-block comprises two 2D convolution layers with residual connections (res-conv-layers) and one attention layer. A group normalization and a SiLU activation follow each res-conv layer. For the up-sample-blocks, we similarly adopt two res-conv-layers followed by group normalization and SiLU activation. Then, a bilinear upsampler is appended in each block, except for the last one. The upsample layers expand the spatial size of noised tokens ($3 \times 8 \times 20\times 20$) to the shape of final triplane grids $(3\times 64\times 320\times 320)$. Taking one feature plane $\mathcal{P}_{XY}$ for instance, the noise tokens $\tokenXY$ with size of ($d_t \times r_X\times r_Y$) are lifted from $d_t$ to $D_P$ in channels and upscaled ($L - 1$) times in the spatial dimensions to the final plane's shape ($ D_P \times R_X\times R_Y$), where $r_X = 20, r_Y=20, d_t=8, D_P=64, R_X = 2^{L-1} \times r_X$ and $R_Y = 2^{L-1} \times r_Y$.

% the hidden dimension of the input noised triplane token is initialized to 16 and then increased to 64 after the triplane generation module. In the triplane generation module, a pre-trained DINOv2 is employed as a scene encoder to produce 768-dimension feature tokens with a patch size of 14. Subsequently, the noised triplane tokens serve as queries for cross-attention with the feature tokens.

\section{Detailed Derivation Formulation in DPA}
Commonly used aggregations introduced entanglement to pose with triplane features. One operation is Hadamard product used in \cite{kplanes}. On account of the multiplicative nature, when gradient update is unstable in the early steps, product among planes causes violent vibration on pose optimization and may bring interference into gradients back propagated from triplane feature, further resulting in updating collision on pose. However, in forward-facing scenes, angles of images or videos are generally consistent, with all objects primarily facing towards cameras. Since scenes emphasize the front view of objects, the frontal features gain prime focus and accordingly only one or two planes may receive a good optimization signal. Meanwhile, unlike the relatively independent update of planes, the parameters of each pose receive optimization signals from all different planes, as the 3D points for feature query are obtained by sampling rays from the corresponding camera.

To disentangle pose with planes, we design our DPA, which is formulated as:
\begin{equation}
\label{eqn:dpa}
\begin{split}
    \mathbf{DPA}(\mathcal{P},\mathbf{x})&=\prod_{k}\Interp\left(\mathcal{P}_k,\Detach(\pi_k(\mathbf{x}))\right)+\sum_{m}\Interp(\Detach(\mathcal{P}_m),\pi_m(\mathbf{x}))\\
    &+\sum_{k\neq m}^{k,m}\Interp(\Detach(\mathcal{P}_k),\Detach(\pi_k(\mathbf{x})))\Interp(\Detach(\mathcal{P}_m),\Detach(\pi_m(\mathbf{x})))+1,\\
\end{split}
% \vspace{-0.3cm}
\end{equation}
where $\Detach(x)$ is the gradient-detached copy of $x$ and $\frac{\partial \Detach(x)}{\partial x}$ is \textit{zero}. Thus, denote that the feature obtained from the proposed aggregation $\hat{F} = \mathbf{DPA}(\mathcal{P},\mathbf{x})$, the core part $\frac{\partial \hat{F}}{\partial \mathbf{x}}$ of pose updating gradients can be expressed as:
\begin{equation}
\label{eqn:novelgrad}
\begin{split}
    \frac{\partial \hat{F}}{\partial \mathbf{x}}&= \frac{\partial \prod_{k}\Interp\left(\mathcal{P}_k,\Detach(\pi_k(\mathbf{x}))\right)}{\partial \mathbf{x}} + \frac{\partial \sum_{m}\Interp(\Detach(\mathcal{P}_m),\pi_m(\mathbf{x}))}{\partial \mathbf{x}}\\
    &+\frac{\partial \sum_{k\neq m}^{k,m}\Interp(\Detach(\mathcal{P}_k),\Detach(\pi_k(\mathbf{x})))\Interp(\Detach(\mathcal{P}_m),\Detach(\pi_m(\mathbf{x})))}{\partial \mathbf{x}} + \frac{1}{\partial \mathbf{x}}\\
    &=\sum_{m}\left(\frac{\partial \Interp\left(\mathcal{P}_m, \Detach(\pi_m(\mathbf{x}))\right)}{\partial \mathbf{x}} \cdot \prod_{k\neq m}\Interp\left(\mathcal{P}_k, \Detach(\pi_k(\mathbf{x}))\right)\right)\\
    &+ \frac{\partial \sum_{m}\Interp(\Detach(\mathcal{P}_m),\pi_m(\mathbf{x}))}{\partial \mathbf{x}}\\
    &+\frac{\partial \sum_{k\neq m}^{k,m}\Interp(\Detach(\mathcal{P}_k),\Detach(\pi_k(\mathbf{x})))\Interp(\Detach(\mathcal{P}_m),\Detach(\pi_m(\mathbf{x})))}{\partial \mathbf{x}} + \frac{1}{\partial \mathbf{x}}.
\end{split}
\end{equation}

Since $\frac{\partial \Detach(x)}{\partial x}$ is \textit{zero} and $\frac{\partial \Interp(\mathcal{P},\Detach(x))}{\partial x}$ is \textit{zero}, we get the final expanded expression of $\frac{\partial \hat{F}}{\partial \mathbf{x}}$ as:
\begin{equation}
    \frac{\partial \hat{F}}{\partial \mathbf{x}} = \sum_{m}\frac{\partial \Interp\left(\Detach(\mathcal{P}_m), \pi_m(\mathbf{x})\right)}{\partial \mathbf{x}},
\end{equation}
where the gradients back propagated to pose from different planes are combined with sum, which is beneficial to pose optimization. Meanwhile, with the gradients for plane $XY$ as:
\begin{equation}
    \frac{\partial \hat{F}}{\partial \mathcal{P}_{XY}} = \frac{\partial \prod_{k}\Interp\left(\mathcal{P}_k,\Detach(\pi_k(\mathbf{x}))\right)}{\partial \mathcal{P}_{XY}},
\end{equation}
and so as the other two planes $YZ,XZ$, our triplane features preserve the expressive ability for scene representation.

\section{More Experimental Details}

\subsubsection{Proposal Sampling and Density Field.} We use a proposal sampling strategy for 3D point sampling and implement it similarly to the one in K-Planes\cite{kplanes}, which is a more compact variant from the proposal sampling strategy in Mip-NeRF 360\cite{mipnerf360}. K-Planes designed a density model with a triplane structure similar to its feature model, and trained it with histogram loss. We borrow the two-stage proposal sampling and basic density models from K-Planes and histogram loss from Mip-NeRF 360 but bond them with the pose-scene joint optimization. Therefore, our density field are forged and updated by another triplane generator. 

\subsubsection{Datasets.}
We conduct experiments on two datasets: LLFF \cite{llff} and NeRF-synthetic dataset \cite{mildenhall2021nerf}. The LLFF \cite{llff} is a real-world dataset consisting of eight forward-facing scenes captured by mobile phones. The NeRF-Synthetic dataset \cite{mildenhall2021nerf} contains pathtraced images of eight objects that exhibit complex geometry and non-Lambertian material.

\subsubsection{Implementation Details.} To achieve the joint optimization of camera poses and triplane, we follow the architecture of K-Planes \cite{kplanes} with some modifications for pose refinement and thus make the joint pose-triplane optimization baseline. Assuming known camera intrinsics, we follow \cite{lin2021barf,wang2021nerfmm} to parameterize camera extrinsics as learnable variables $T \in SE(3)$, where the rotations are optimized in axis-angle $\phi_i \in \mathfrak{s o}(3)$. 

In the first stage, we utilize two separate Adam optimizers to independently optimize the triplane generator and camera poses. Specifically, the learning rate for the generator linearly increases from 0 to 0.002 in the first 128 steps of training and decreases with cosine-annealing until the second stage. The learning rate for the camera pose is set to 0.001. We switch to the second stage of learning after 4000 steps with our proposed warm-start strategy. 

In the second stage, we discard the triplane generator and set the generated triplane as learnable variables, utilizing a new Adam optimizer for optimization. The learning rates for triplane and camera parameters linearly increase from 0 to 0.03 and 0.001 respectively within the first 128 steps from the second stage, ensuring a smooth transition to the second-stage direct optimization approach. Similar to the first stage, these learning rates decrease exponentially to $1 \times 10 ^{-5}$ in the remaining steps. 

In both stages, we randomly sample 4096-pixel rays at each optimization step. Following \cite{kplanes}, we employ proposal sampling to sample 48 points along each ray for subsequent volume rendering. For forward-facing LLFF, we utilize normalized device coordinates (NDC) to better allocate our resolution and enable unbounded depth. During the evaluation, we follow \cite{lin2021barf} to run additional steps of test-time learning on the frozen trained models and only optimize poses for testing during this procedure. Our model is implemented with PyTorch and trained 60k (for NeRF-synthetic \cite{mildenhall2021nerf} and 70k for LLFF\cite{llff}) epochs per scene on an NVIDIA Tesla V100 GPU.

\subsubsection{Evaluation Criteria.}
For novel view synthesis evaluation, we first conduct test-time optimization on each testing pose as proposed in \cite{lin2021barf} to eliminate minor errors caused by the misalignment of the test phase and training phase camera poses. We report PSNR, SSIM \cite{ssim} for novel view synthesis evaluation. For camera pose evaluation, we follow previous works \cite{lin2021barf,l2gnerf} to perform Procrustes analysis for aligning the training poses and the GT poses before calculating the rotation error (in degree) and translation error (scaled by 100).

\section{More Experimental Results}

\subsubsection{Additional Qualitative Results of Novel View Synthesis.} 
\begin{figure}[!ht]
    \includegraphics[width=0.95\linewidth]{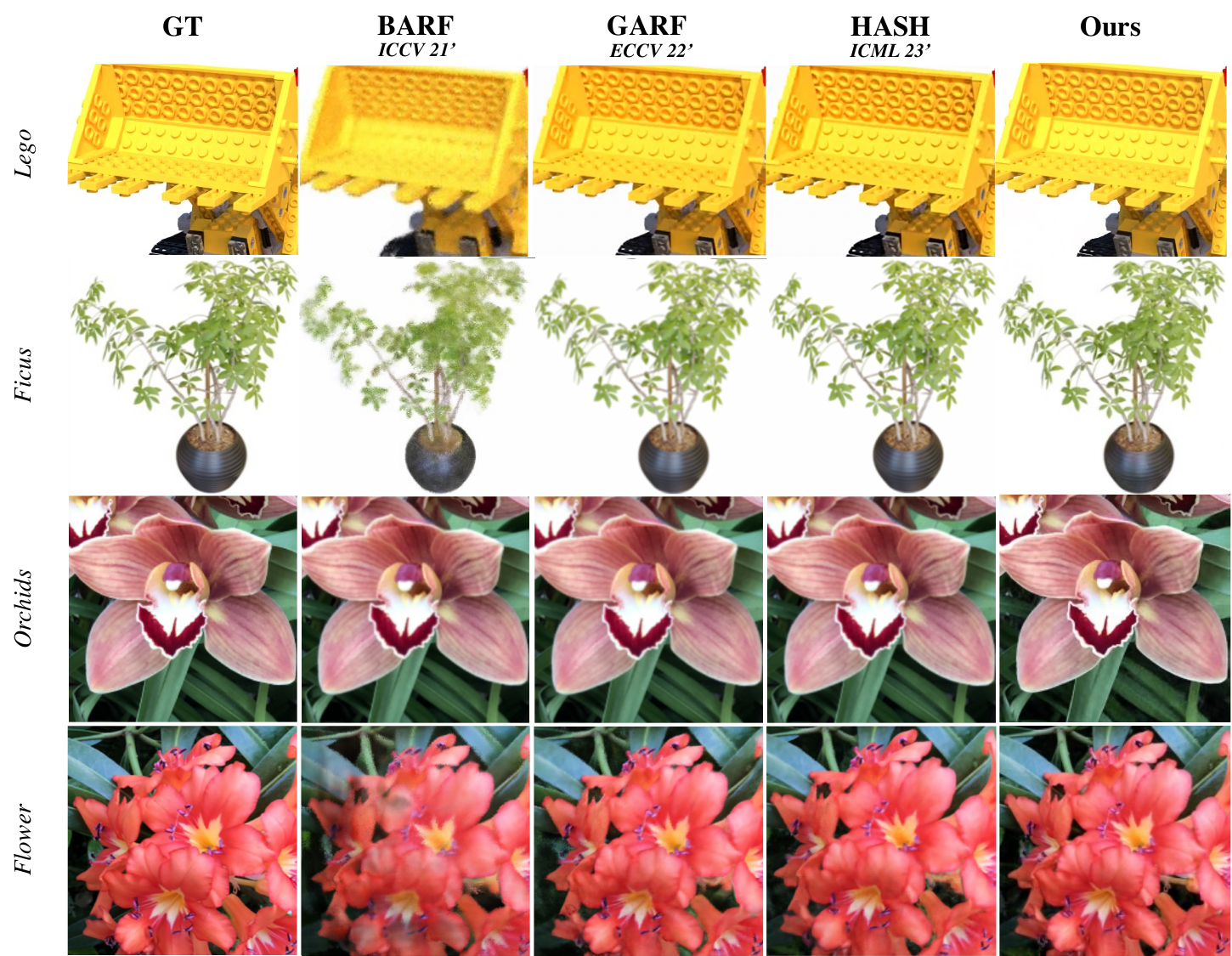}
    \centering
    \caption{
    \textbf{Additional qualitative results of novel view synthesis.}  
    } 
    \label{fig:suppres3}
\end{figure}

We provide additional novel view synthesis comparisons as shown in \cref{fig:suppres3}. Since the implementation of GARF \cite{chng2022garf} and HASH \cite{robust_hash} are unavailable, we directly use the results reported in their paper for comparison.

\subsubsection{Ablation on Scene Texture Embedding.}

As shown in \cref{tab:abl}, we additionally perform an ablation of the scene texture embedding module in the triplane generator. The results show a degradation in the quality of the novel view rendering after removing this module from our model. This demonstrates that our Scene Texture Embedding module introduces more scene texture prior for triplane generation, thus enhancing the triplane representation.
\begin{table}[]
\label{tab:abl}
\centering
\caption{Ablations on applying the Scene Texture Embedding in the triplane generator in real-world LLFF dataset \cite{llff}.}
\begin{tabular}{ccc}
\hline
Settings                                                   & PSNR$\uparrow$ & SSIM$\uparrow$ \\ \hline
\multicolumn{1}{c|}{Ours with Scene Texture Embedding}    & 25.90          & 0.828          \\
\multicolumn{1}{c|}{Ours without Scene Texture Embedding} & 25.35          & 0.813          \\ \hline
\end{tabular}
\end{table}

Besides, we further provide results across varied scenes in Table below, the proposed STE improves a minimum of 0.18 db on \textit{Lego} and a maximum of 1.23 db on \textit{Fortress}, indicating consistent improvements. 

\begin{table}[h!]
\centering
\caption{More detailed ablations on the Scene Texture Embedding module.}
\resizebox{\columnwidth}{!}{
\begin{tabular}{l|cccccc}
\hline
                 & Fern          & Fortress      & Room          & Flower        & Lego          & Drums         \\ \hline
Ours without STE & 24.68& 29.56         & 32.84         & 26.67         & 32.01         & 25.73         \\ 
 Ours             &  
 25.70 (\textcolor{OliveGreen}{\textbf{+1.02}})& 30.79 (\textcolor{OliveGreen}{\textbf{+1.23}}) & 33.95 (\textcolor{OliveGreen}{\textbf{+1.11}}) & 27.06 (\textcolor{OliveGreen}{\textbf{+0.39}}) & 32.19 (\textcolor{OliveGreen}{\textbf{+0.18}}) &26.10 (\textcolor{OliveGreen}{\textbf{+0.37}}) \\ \hline
\end{tabular}
}
\end{table}

\subsubsection{More Empirical Experiments on Inappropriate Learning Signals from Different Feature Grids.}
\begin{figure}[!ht]
  \centering
  \includegraphics[width=\linewidth]{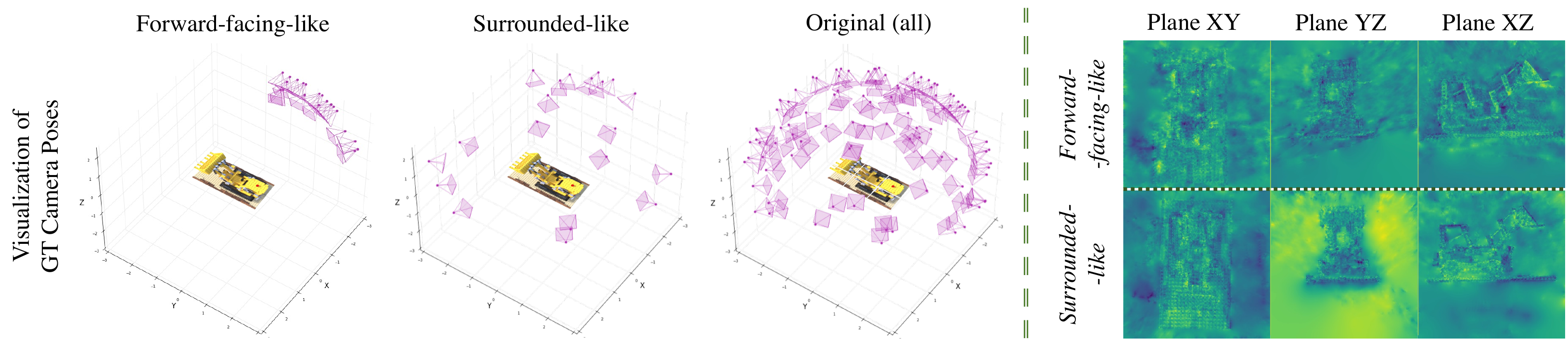}
  % \vspace{-5mm}
   \caption{Visualization of camera distribution and plane features.}
   \label{fig:R2_vis}
   % \vspace{-0.5cm}
\end{figure}

We further provide empirical experiments from \textit{Lego} as shown in \cref{fig:R2_vis}. The `forward-facing-like' cameras are gathered within a limited view angle towards PlaneXZ and the `surrounded-like' set almost covers the whole scene. Triplane receives less supervision (PlaneXY and PlaneYZ) in `forward-facing-like' scenes with more noisy and incomplete feature textures.

\subsubsection{Additional Ablations on Two-Stage Warm-Start Training.}
\begin{figure}[!ht]
  \centering
  \includegraphics[width=0.6\linewidth]{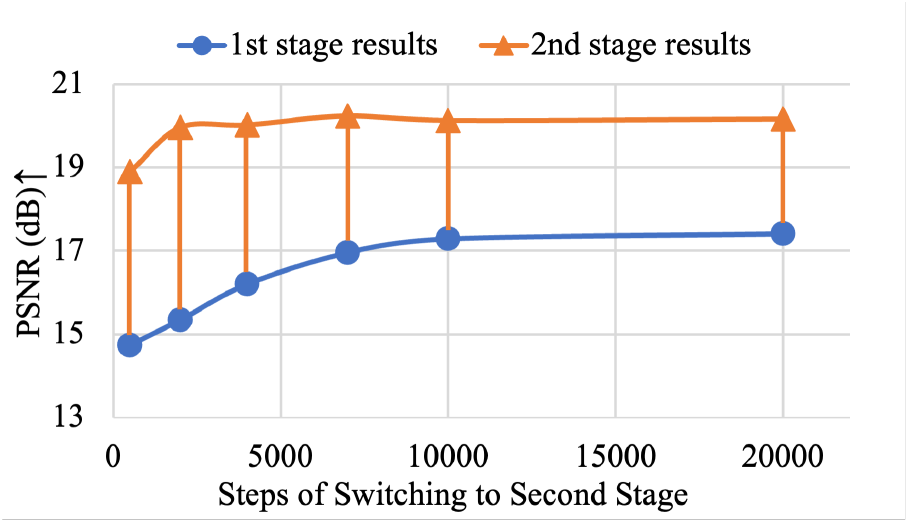}
  % \vspace{-5mm}
   \caption{Comparisons of different steps of switching to second stage.}
   \label{fig:R4_steps}
   % \vspace{-0.5cm}
\end{figure}

We perform a two-stage system ablation experiment on the challenging scene \textit{orchids} by switching from the first stage to the second stage at different training steps as shown in \cref{fig:R4_steps}. We can see our two-stage strategy (\textcolor{orange}{orange}) stably improves the performance regardless of the quality of the first stage (\textcolor{blue}{blue}). 

\subsubsection{Comparison of Utilizing Noise Tokens and Image Tokens as Input.}
\begin{figure}[!ht]
  \centering
  \includegraphics[width=0.9\linewidth]{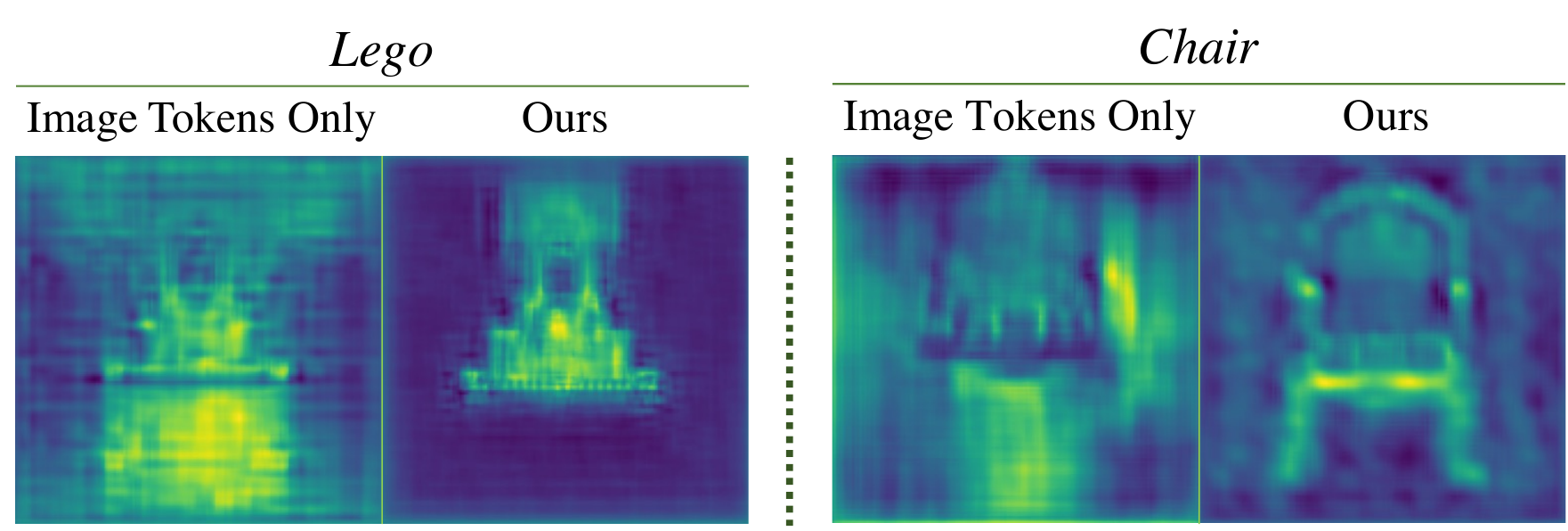}
  % \vspace{-5mm}
   \caption{Visual comparisons of feature plane between only image feature tokens inputs and our full inputs with noise tokens at 1000 training steps.}
   \label{fig:R4_vis_v1}
   % \vspace{-0.5cm}
\end{figure}

We initialize fixed triplane noise tokens to introduce \textit{spatial priors}. We provide more comparison as shown in \cref{tab: comp_imagetoken} and \cref{fig:R4_vis_v1}.

\begin{table}[!ht]
\centering
\caption{Qualitative comparison between using image tokens and noise tokens as input.}
\resizebox{0.7\linewidth}{!}{
\begin{tabular}{l|cccc} 
\hline
Scenes & Room & Fern & Lego & Chair \\ \hline
Image Tokens Only & 31.90 & 24.48 & 30.07 & 34.27 \\
Ours & \textbf{33.95} & \textbf{25.70} & \textbf{32.19} & \textbf{37.78} \\ \hline

\end{tabular}

}
\label{tab: comp_imagetoken}
\end{table}

\subsubsection{Visual Comparisons with Baseline Joint Estimation.}
\begin{figure}[!ht]
    \includegraphics[width=0.95\linewidth]{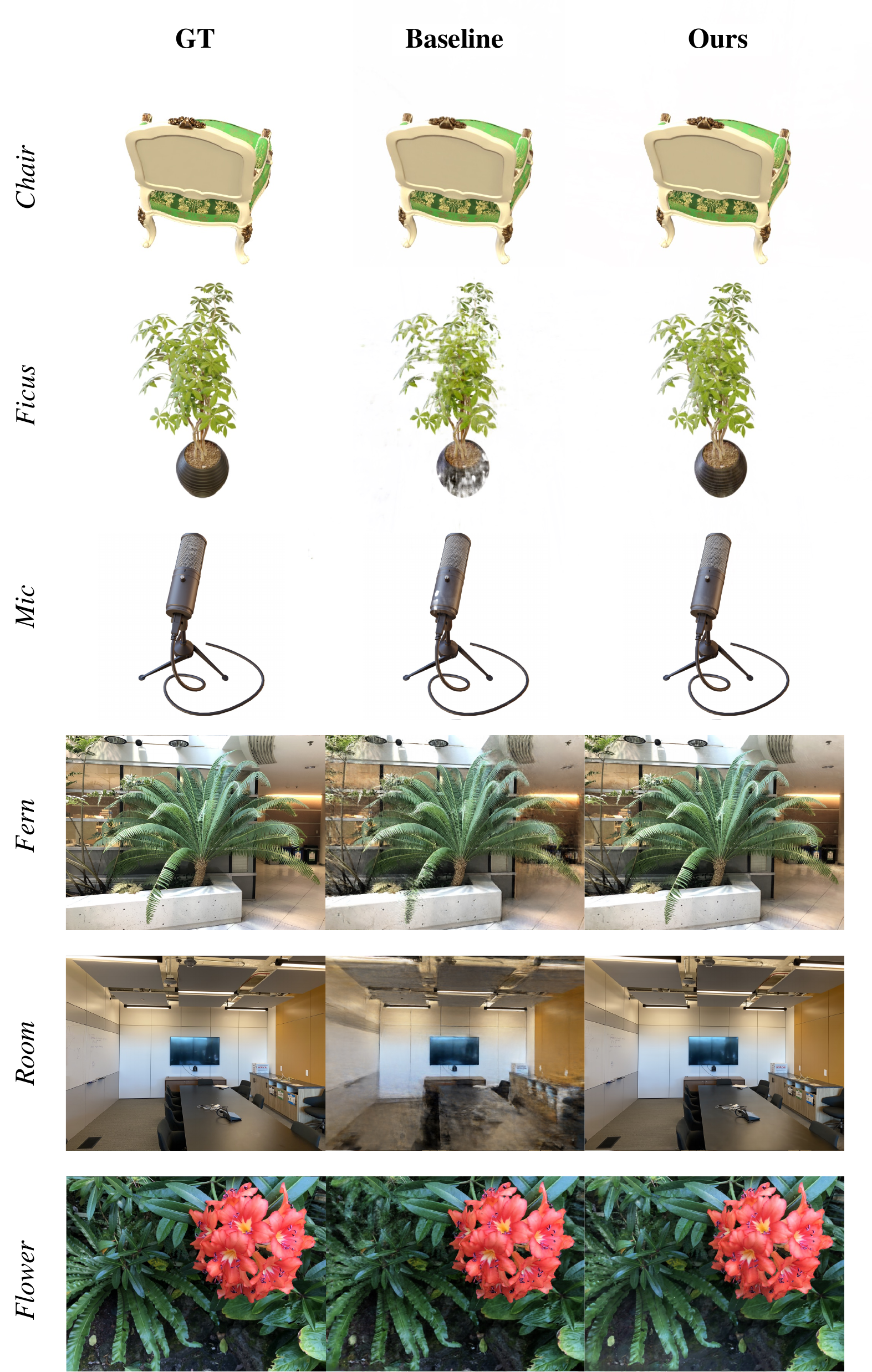}
    \centering
    \caption{
    \textbf{Visual comparisons between our full model and the baseline direct joint pose-triplane optimization.}  
    } 
    \label{fig:suppres}
\end{figure}

We provide additional visualization comparisons as shown in \cref{fig:suppres} to illustrate the effectiveness of our approach. Compared to the baseline simple combination of pose estimation and triplane-NeRF optimization, our approach achieves higher-quality visual effects, which demonstrates that our proposed method mitigates the errors caused by local updating and entanglement, leading to better pose estimation and novel view rendering results.

\subsubsection{Limitations and Future Works.}Although our method can recover the camera pose and triplane radiance fields effectively, there is still room for improvement. In the future, we will explore introducing more powerful 3D representations such as 3D Gaussian Splatting \cite{3dgs} into the joint optimization pipeline and try to further reduce the dependence on camera pose initialization.

\end{document}